\documentclass{article}

\usepackage[preprint]{corl_2021} % Uncomment for pre-prints (e.g., arxiv); This is like ``final'', but will remove the CORL footnote.

\usepackage[utf8]{inputenc}
\usepackage{graphicx}
\usepackage{appendix}
\usepackage{pdfpages}
\usepackage{anyfontsize}
\usepackage{setspace}
\usepackage{amsmath}
\usepackage[font=footnotesize]{caption}
\usepackage{todonotes}
\usepackage{amssymb}
\usepackage{booktabs}
\usepackage[normalem]{ulem}
\useunder{\uline}{\ul}{}
\usepackage{wrapfig}
\usepackage{multirow}
\usepackage[bottom]{footmisc} % put footnote under figures

\definecolor{red}{rgb}{1.00,0.00,0.00}
\definecolor{blue}{rgb}{0.00,0.00,1.00}

\title{\Large Lifelong 3D Object Recognition and Grasp Synthesis \\using Dual Memory Recurrent Self-Organization Networks}

\author{
  Krishnakumar Santhakumar\\
  Department of Artificial Intelligence\\
  University of Groningen, Netherlands\\
  \texttt{k.santhakumar@student.rug.nl} \\
  \And
   Hamidreza Kasaei \\
   Department of Artificial Intelligence\\
   University of Groningen, Netherlands\\
   \texttt{hamidreza.kasaei@rug.nl} \\
}

\begin{document}
\maketitle

%===============================================================================

\begin{abstract}
 Humans learn to recognize and manipulate new objects in lifelong settings without forgetting the previously gained knowledge under non-stationary and sequential conditions. In autonomous systems, the agents also need to mitigate similar behavior to continually learn the new object categories and adapt to new environments. In most conventional deep neural networks, this is not possible due to the problem of catastrophic forgetting, where the newly gained knowledge overwrites existing representations. Furthermore, most state-of-the-art models excel either in recognizing the objects or in grasp prediction, while both tasks use visual input. The combined architecture to tackle both tasks is very limited. In this paper, we proposed a hybrid model architecture consists of a dynamically growing dual-memory recurrent neural network (GDM) and an autoencoder to tackle object recognition and grasping simultaneously. The autoencoder network is responsible to extract a compact representation for a given object, which serves as input for the GDM learning, and is responsible to predict pixel-wise antipodal grasp configurations. The GDM part is designed to recognize the object in both instances and categories levels. We address the problem of catastrophic forgetting using the intrinsic memory replay, where the episodic memory periodically replays the neural activation trajectories in the absence of external sensory information. To extensively evaluate the proposed model in a lifelong setting, we generate a synthetic dataset due to lack of sequential 3D objects dataset. Experiment results demonstrated that the proposed model can learn both object representation and grasping simultaneously in continual learning scenarios. %\cred{Experimental videos and supplementary material are available at: \url{https://youtu.be/4Bc9tjzbxDQ}, \url{https://github.com/krishkribo/3D_GDM-RSON}}.
 
\end{abstract}


%===============================================================================
\section{Introduction}

Robots need to learn a set of perceptual and manipulation skills to perform complex tasks in dynamic environments. 
As an example, consider cleaning a table task.  A robot needs to know which objects exist in the collection, where they are, and how to grasp and manipulate various objects. In such scenarios, a robot may also face never-seen-before objects. Therefore, such robots need to learn new information's overtime as it is not possible to pre-program everything in advance. The ability to learn new knowledge in the environment while retaining the previously acquired knowledge is referred to as \textit{continual} or \textit{lifelong} learning. 
%Continual Learning is one of the key elements which enables robots to learn in lifelong settings. 
%The problem of lifelong learning is the long-standing challenge in robotics, machine learning, and neural networks~\cite{french1999catastrophic}\cite{hassabis2017neuroscience}. 

In the deep learning era, most robots use deep learning models to learn perceptual and manipulation skills. The conventional deep learning models are trained on the fixed batches of the large datasets, which is not suitable for continual learning. In continual learning settings, training data become progressively available over time. Therefore, the model developed for continual learning tasks should need to adapt to the newly introduced categories. One of the major problems that needs to be addressed in such settings is \textit{catastrophic forgetting} which is caused due to periodic decrease of stability-plasticity dilemma~\cite{10.3389/fpsyg.2013.00504}. 
When a conventional deep learning model is trained on sequential tasks, the performance of the network on previously learned tasks is reduced~\cite{kemker2018measuring} \cite{maltoni2019continuous}, and retraining from scratch needs to happen. Such a retraining procedure is computationally expensive and required large memory for storing all the encountered data. Furthermore, in most lifelong learning scenarios, the direct access to the previous experiences is restricted~\cite{thrun1995lifelong}. Therefore, conventional deep learning methods are not applicable in lifelong robot learning applications, as they need to learn the new object categories on-site and fast without forgetting the previously learned tasks. 

The complementary learning systems (CLS) theory is the most likely path to break these barriers. In particular, CLS provides the computational framework basics for modeling memory consolidation and retrieval \cite{mcclelland1995there}. The CLS system holds two interdependent operations, namely recollecting the separate episodic events and learning the statistical structure, which are mediated by the interplay providing means for \textit{episodic memory} and \textit{semantic memory} \cite{mcclelland1995there}\cite{kumaran2016learning}. 

\begin{figure}
    \centering
    \includegraphics[width=\linewidth]{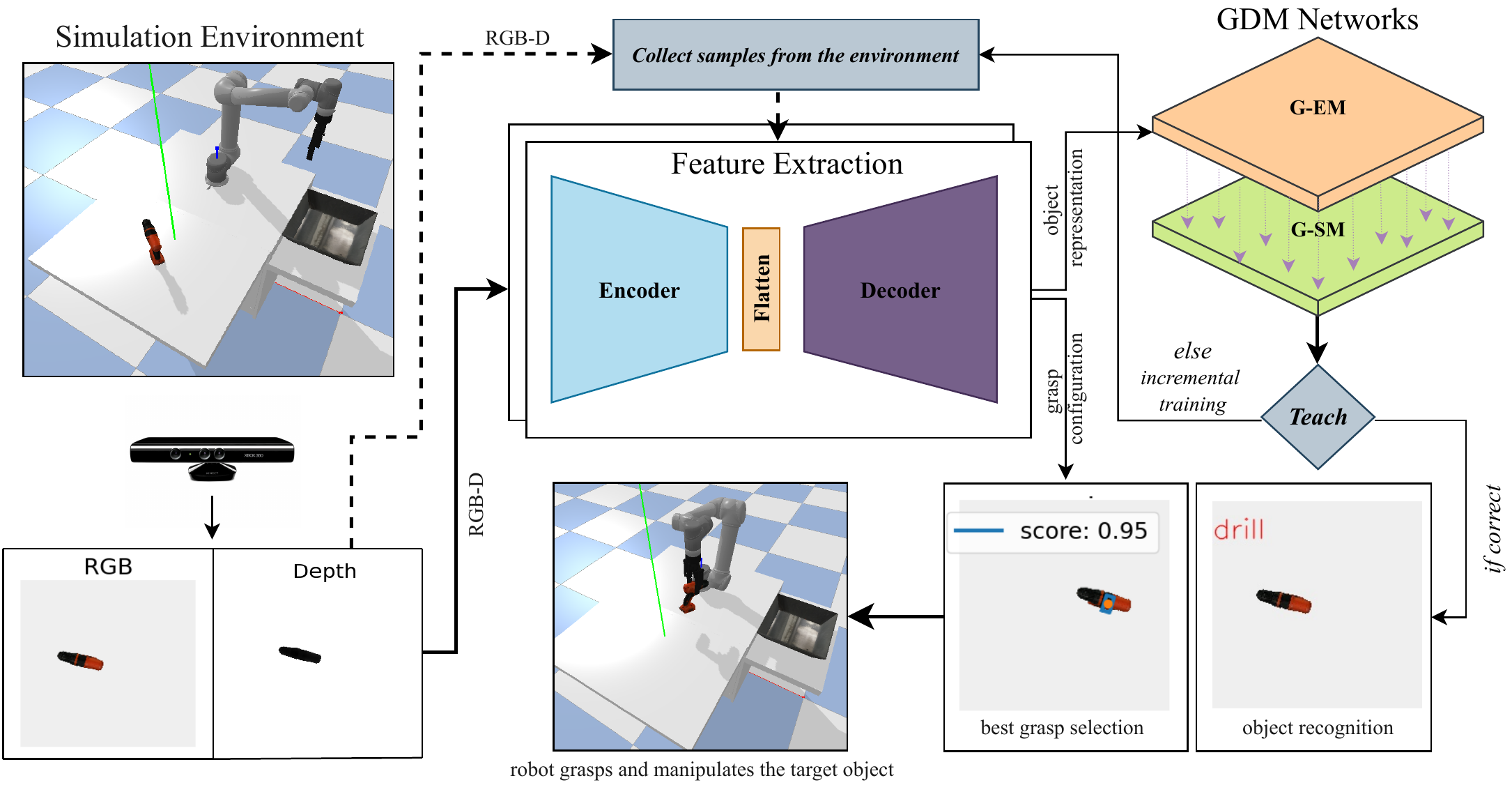}
    \caption{Overall system architecture for continuous object recognition and grasping. RGB-D data representing the 3D object in the environment is fed to generative autoencoder to obtain, \textit{(i)} the object representation for object recognition using GDM networks and \textit{(ii)} grasp configuration for object grasping. If the new object category or instance is introduced to the robot, based on the teacher input the system automatically collect the samples from the environment and its associated label information to continually learn and adapt to the new sensory information.}
    \label{fig:1}
\end{figure}

In this paper, we propose a hybrid continual learning model for object recognition and grasp synthesis (see Fig.~\ref{fig:1}).  {To the best of our knowledge, our work is the first effort to address object recognition and grasping simultaneously using dual memory recurrent self-organization networks in continual learning fashions.} The proposed architecture consists of Growing Dual-Memory (GDM) recurrent self-organization networks~\cite{GDM/10.3389/fnbot.2018.00078}. {The underlying reason for selecting the GDM network over other possible architectures is that the GDM network consists of a dual memory structure (\textit{episodic memory} and \textit{semantic memory}) that dynamically adapts the number of neurons and synapses. This structure makes it possible to not only learn the spatiotemporal relations from the input data (episodic memory) but also to learn high-level compact representations from the learned temporal representations (semantic memory). Additionally, the associative matrix labeling in the GDM network enables the model to perform instance and category level classification in an unsupervised fashion.  Furthermore, the GDM networks are able to handle the number of missing labels to a great extent without catastrophic forgetting. More specifically, to learn the representation of the sensory input, the episodic memory learns in an unsupervised fashion to dynamically adapt the obtained object representation at the \textit{instances level}, whereas semantic memory is responsible to learn a compact representation of the statistical regularities embedded in the episodic memory at the \textit{category level}.} In our approach, the neural activity pattern embedded in episodic memory is periodically replayed to the episodic and semantic memory, using pseudo-rehearsal or intrinsic memory replay  \cite{robins1995catastrophic}, to mitigate the catastrophic forgetting. Figure~\ref{fig:1} shows an overview of the proposed approach.

%for the lifelong learning of the spatio-temporal representations, and a convolutional neural network (CNN) based autoencoder. {The GDM network comprises two recurrent self-organization memories (\textit{episodic memory} and \textit{semantic memory}) that dynamically adapt the number of neurons and synapses. In particular, to learn the representation of the sensory input, the episodic memory learns in an unsupervised fashion to dynamically adapt the obtained object representation at the \textit{instances level}, whereas semantic memory is responsible to learn a compact representation of the statistical regularities embedded in the episodic memory at the \textit{category level}.} In our approach, the neural activity pattern embedded in episodic memory is periodically replayed to the episodic and semantic memory, using pseudo-rehearsal or intrinsic memory replay  \cite{robins1995catastrophic}, to mitigate the catastrophic forgetting. Figure~\ref{fig:1} shows an overview of the proposed approach.

{In summary, the key contributions of this paper are: (\textit{i}) a new hybrid end-to-end system which is capable of continual learning of object categories and grasp synthesis;  (\textit{ii}) we modified the GDM network to be suited for lifelong 3D object recognition and also the GR-ConvNet model to improve object grasping performance; (\textit{iii}) we conducted extensive sets of experiments in batch learning and the incremental learning scenarios; (\textit{iv}) we also evaluate the proposed 3D object recognition and grasping in a simulation robotic settings.}

The remainder of this paper is organized as follows: In Section~\ref{related_work}, the related works are reviewed. The detailed methodology of our approach is explained in Section~\ref{methods}. Experimental results is then discussed in Section~\ref{experiments_results}. Finally, the conclusion and the possible future direction are explained in Section~\ref{conclusion_future_work}.

%%%==========================================================

\section{Related Work}
\label{related_work}
Although an in-depth review is beyond the scope of this work, we discuss a few recent efforts in continual learning, object recognition, and grasping.

Several works have been published to address continual learning using different techniques, such as complementary learning systems (CLS)~\cite{mcclelland1995there}, regularization methods ~\cite{kirkpatrick2017overcoming, zenke2017continual, fernando2017pathnet}, dynamical architectures ~\cite{yoon2017lifelong, MARSLAND20021041, gepperth2016bio, part2016incremental, parisi2017lifelong}, and memory replay techniques \cite{Delange_2021}. 
%The CLS theory provides the basic foundation for the computational framework which aims to generalize new sensory experiences while retaining the specific memories in a lifelong learning fashion. 
We refer the reader to a brief review on continual learning in deep network by Parisi et al. \cite{PARISI201954}. It provides a good insights into the continual learning using different approaches. 

In general, the problem of continual learning is usually addressed by several regularization methods. The elastic weight consolidation (EWC)~\cite{kirkpatrick2017overcoming} addresses continual learning in supervised and reinforcement learning (RL) scenarios by mitigating the catastrophic forgetting. Zenke et al., \cite{zenke2017continual} proposed an approach to alleviate the catastrophic forgetting by allowing the individual synapses in the learning model to estimate their importance in the learned task. 
An ensemble method, named PathNet~\cite{fernando2017pathnet}, uses a genetic algorithm to find the optimal path through a neural network of fixed size to find which parts of the neural network can be reused for learning new task while freezing task-relevant parts is developed to avoid catastrophic forgetting. Although such regularization methods provide a way to alleviate the problem of catastrophic forgetting, they are limited by the number of neural resources for learning new tasks over time which may lead to the performance trade-off between the old task and the new task. 

To address the limitation of neural representations in the regularization methods, several dynamical architectures were proposed. For instance, Dynamically Expanding Network (DEN)~\cite{yoon2017lifelong} increases the trainable parameters while learning the new task using network expansion and selective retraining by sparse regularization in a supervised learning paradigm. Such networks learn the representation of new tasks by dynamically increasing the network size. This way, the problem of limited neural resource is addressed. Recently, S. Jain \& H. Kasaei \cite{jain20213dden} reported state-of-the-art results in open-ended 3D object recognition using pre-trained convolutional neural network (CNN) and DEN. Similar to DEN, the combination of the self-organizing incremental neural network and a pre-trained CNN, proposed by Part and Lemon~\cite{part2016incremental,part2017incremental}, allows the network to grow in a continuous object recognition scenarios. Although the dynamically expandable networks adapt to learn a new task, it needs to have access to the entire dataset while learning the new task which increases the storage complexity.

The continual learning models for robotics application have to address the problem of complex data storage. Towards this goal, Marsland et al.,~\cite{MARSLAND20021041} proposed a neural network, which grows when required (GWR) based on the synaptic neural activity triggered by the input data distribution to the best matching similarity nodes in the network, and it also deals with the dynamic data distributions. An extended version of the GWR model, called Gamma-GWR~\cite{parisi2017lifelong}, embeds the gamma memory \cite{principe1994analysis} during the neuron growth for learning the short-term temporal relation representations of the input data distribution in the absence of external sensor information. This extended version of GWR network address the problem of data storage in continual learning by learning the short-term temporal relation. Using Gamma-GWR, Parisi et al., \cite{parisi2017lifelong} showed the state-of-the-art results in batch learning scenarios with missing and corrupted sample labels. 

Apart from the above-mentioned methods, the concept of dual-memory systems were developed to address short-term and long-term memory consolidation. The system in which each synaptic connection has two weights: the plastic weights (to preserve long-term knowledge) and the fast-changing weights (which holds the temporal short-term knowledge). One of such dual-memory system was proposed by Gepperth and Karaoguz \cite{gepperth2016bio}, using modified self-organizing maps (SOM) and SOM extended with short-term memory (STM) to address the incremental learning task by alleviating the catastrophic forgetting. In their work, they used STM to store the previously learned knowledge and replayed back while learning the new task. Even though SOM $+$ STM address continual learning it also has certain limitations. Since the STM has the limited capacity it overwrites the old task while learning a new task and it also requires storing the entire dataset during incremental training.

The above-mentioned methods designed for the classification of the static data representations in supervise learning paradigm. In more natural settings, the data representations are sequential, where the underlying spatio-temporal relations are incrementally available over time (i.e. objects information with different features representations). For the continual learning, a sequential dataset with temporal meaningful relations needs to be used.

%In our paper, we focused on the learning algorithm which dynamically adapts to the input data distribution, with a memory replay technique to preserve the previously gained knowledge while learning the new tasks without storing the entire training dataset. 

The growing dual-memory (GDM) network architecture~\cite{GDM/10.3389/fnbot.2018.00078} is better at representing the spatio-temporal relation of the input data in the lifelong settings with reduced storage and computational complexity. Furthermore, the GDM network achieved state-of-the-art result in continuous object recognition scenarios. The GDM network consists of deep transfer learning-based pre-trained CNN, followed by the two different recurrent self-organizing gamma grow when required networks (Gamma-GWR), named episodic memory (learns the sensory experience) and the semantic memory (learns the task-relevant signals). The GDM memories can dynamically adapt the number of neurons and synapses based on the input data distribution. Using the intrinsic memory replay or pseudo-rehearsal the problem of catastrophic forgetting is alleviated, in which the previous memories are revisited without storing all the data samples by the period replay of the previously learned temporal synapses (during training) \cite{robins1995catastrophic}. In our approach, we utilized the GDM networks to continually learn the objects categories in an open-ended scenario and, at the same time, we enhanced its prediction performance by introducing the regulated neuron removal based on the neuron activity and its knowledge about the input data. Moreover, We employed different similarity measures to obtain the optimal performance with the input data to estimated the best matching unit (BMU) in the GDM networks. 

%Our work also focuses on the efficient way of object grasp prediction based on the point cloud representation of the input data. 

Several deep learning based algorithms have been proposed for object grasping recently. For examples, Morrison et al., proposed a generative grasping convolutional neural network (GG-CNN)~\cite{morrison2018closing} that can predict pixel-wise grasp configuration for never-seen-before objects. Another approach, named Res-U-Net~\cite{li2020learning}, used encoder-decoder based CNN architecture to predict objects' grasp affordances first, and then used a search policy to find the best path to approach and grasp the target object. In another work, Kasaei et al.,~\cite{kasaei2021mvgrasp} proposed a method to address multi-view 3D object grasping based on convolutional auto-encoder. Kumra et al.,~\cite{kumra2021antipodal} developed a generative residual convolutional neural network (GR-CovNet) which generates the antipodal grasp for the given n-channel input. 
%This system consists of five residual blocks in the encoder architecture which results in achieving the state-of-the-art accuracies of about 97.7\% and 94.6\% on the Cornell and Jacquard \cite{8593950} grasping dataset. Furthermore, they achieved a good grasping success rate on the household and adversarial objects. 
All the above reviewed methods are good at predicting the grasp synthesis for the given 3D objects but they are not able to recognize the object category label simultaneously. To address the simultaneous object category recognition and grasping, 
{ Asif et al.,~\cite{asif2017rgb} considered object grasping and recognition as two independent tasks. In particular,  they formulated object recognition as a batch learning task and object grasping as a heuristic (handcrafted) approach. More specifically, they assumed that all the categories are known in advance and the robot is trained based on hierarchical cascaded forests. Therefore, the knowledge of the robot is fixed after the training phase and, robot re-programming is needed by any changes in the environment. These types of approaches work well in factory-like domains, where everything is predefined, but they are too fragile to be used in a dynamic environment, where the number of categories is not known in advance and the training instances should be extracted from online/onsite experiments.}
Towards addressing this limitation, Kasaei et al.,\cite{kasaei2021simultaneous} coupled a generative mixed auto-encoder with a probabilistic 3D object recognition approach to simultaneously predict the pixel-wise grasp configuration and recognize the objects in open-ended domains. The model receives the input from multiple views of the given 3D object, and then, simultaneously predicts the pixel-wise grasp synthesis and a compact representation for an active object recognition task. In particular, the authors proposed an active learning strategy to teach, ask, or correct the prediction of the model while learning the object categories in an open-ended scenario. Unlike our work, these approaches followed a single-shot prediction and completely discard the spatio-temporal information.   

%Apart from the above-mentioned methods, which primarily focus on predicting the object grasping. We are interested in the network which servers as a feature extractor with considerable output dimensions for object recognition learning, and also to simultaneously predict the grasp representation for the given 3D object.  

%%%%%%%%%%%%%%%%%%%%%%%%%%%%%%%%%%%%%%%%%%%%%%%%%%%%%
%%%%%%%%%%%%%%%%%%%%%%%%%%%%%%%%%%%%%%%%%%%%%%%%%%%%%
\section{Methods}
\label{methods}
\begin{figure}[!t]
    \centering
    \includegraphics[width=\linewidth]{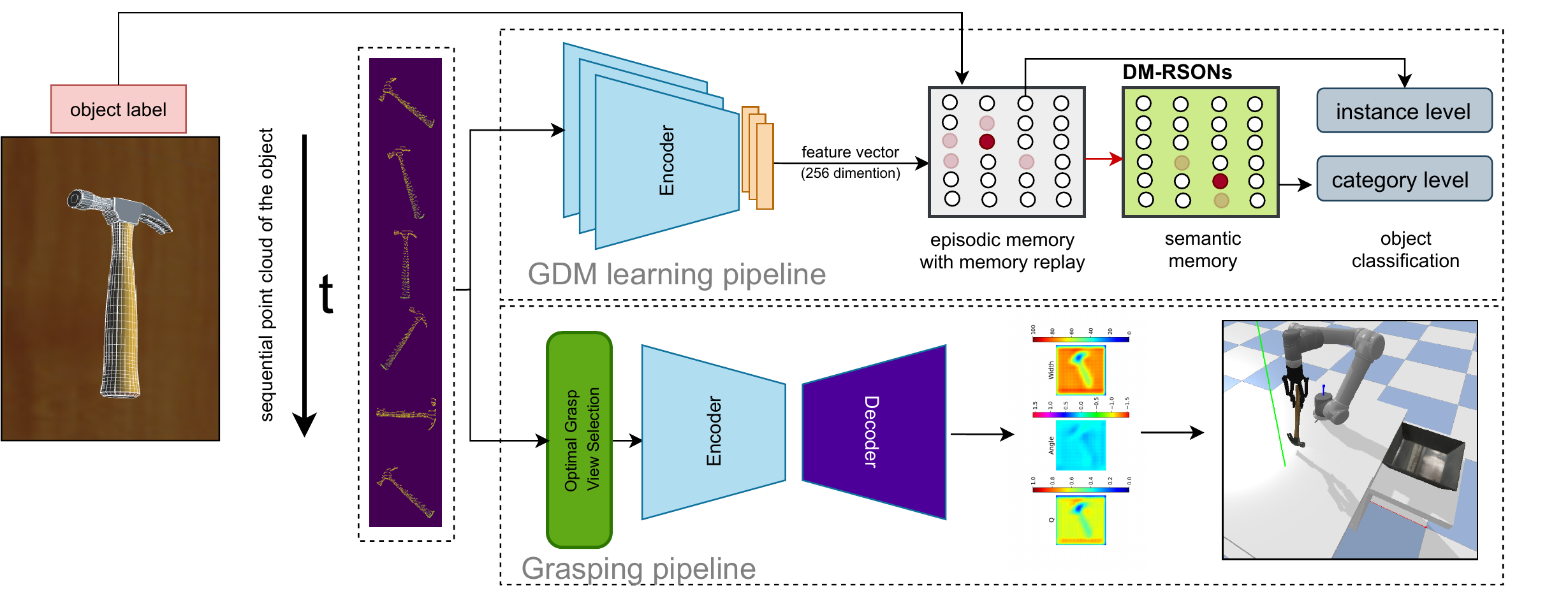}
    \caption{Proposed model architecture: for continual object recognition and grasp synthesis learning using mixed auto-encoder, and growing dual-memory network (GDM). The sequential point cloud samples from the input 3D object are initially generated and converted to an RGB-D image, then fed to the network to obtain instance and category level object representation and pixel-wise grasp configuration. For object recognition, all the sequential samples are being used and for the grasping, the object view with maximum entropy is used.}
    \label{fig:2}
\end{figure}

Our approach to lifelong learning of object recognition and grasp synthesis comprises two main components: (\textit{i}) an autoencoder model is developed to extract a compact feature vector ($256$ dimensions) that is used for object recognition purposes as well as pixel-wise grasp prediction (see Fig.~\ref{fig:3}); (\textit{ii}) a recurrent GDM network, consisting of episodic and semantic memory, to predict the instance and category level. Figure \ref{fig:2} shows the overall architecture of the proposed model.   

Initially, the system receives the multi-view representation of the sequential 3D point cloud data for the object at different time frames \textit{t}. The obtained point cloud is then converted into the depth image serves as an input of the network, $x^i_{t} \in \mathbb{R}^{W \times H}$, where $W$ and $H$ represent the width and the height of the image respectively. In our approach, the RGB-D data with time-dependent sequential input is passed to the encoder to get the $256$ dimensional feature vector. The Gamma-GWR uses distance measures as a metric to compute best matching units (BMUs). To eliminate the discrimination caused by high dimensional and spare data representation, we performed a convolution operation to reduce the dimension of the feature vector to $256$. The obtained representation is then used as an input to the growing dual memory recurrent self-organizing networks to classify the object categories. In parallel to the object category prediction, the sequential RGB-D data is then fed to an entropy-based optimal view selection function. At the time $t=1$, the optimal view will be equal to the input data.

The best view is then passed to the network to generate a pixel-wise antipodal grasp configuration map, $\textbf{G}$, in the form of quality, ($\textbf{q} \in \mathbb{R}^{W \times H}$), angle ($\mathbb{\phi} \in \mathbb{R}^{W \times H}$), and width $\textbf{d} \in \mathbb{R}^{W \times H}$ maps. The point with the highest grasp quality is then selected and converted to robot coordinate to be executed by the robot, i.e., $(u,v) \leftarrow \operatorname{g^*} = \operatorname*{argmax}_\mathbf{q} ~ \mathbf{G}$. 

{Note that our approach differs from imitation learning, where the model or agent mimics the state and actions of an external user or agent. In our approach, the model learns by exploring the sensory information (i.e., images and point cloud data) and not by mimicking the external agent or user. The role of user in our approach is to regulate the learning by either correcting the misclassifications, or by introducing new instances or categories. The GDM network grows and learns in unsupervised learning fashion which is similar to exploration and exploitation dilemma.}

\subsection{Object representation and grasping}

\begin{figure}[!t]
    \centering
    \includegraphics[width=\textwidth]{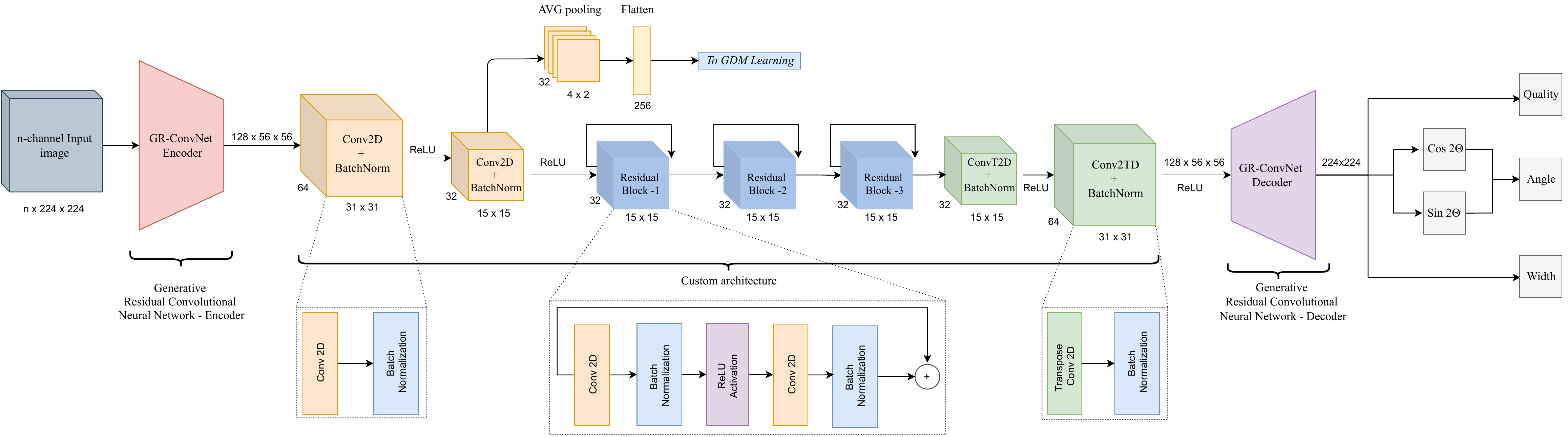}
    \caption{Proposed autoencoder architecture: \textit{(i)} to extract feature vector for object recognition learning, and  \textit{(ii)} to predict the quality, angle, and width images (which then used to estimate the grasp points), for the given n-dimensional RGB-D image input.}
    \label{fig:3}
\end{figure}

We select the generative residual convolutional model (GR-ConvNet)~\cite{kumra2021antipodal} as the backbone of our architecture, since GR-ConvNet showed the state-of-the-art results in object grasping. In particular, we modified the architecture of the GR-ConvNet to be used for both object recognition as well as object grasping. 
%We fine-tuned the layers of pre-trained GR-ConvNet to adapt to the newly added convolution layers.
Figure \ref{fig:3} shows the overall architecture of the proposed autoencoder architecture. 

The output of second convolution layer in the encoder part of the network is used as an object representation. It should be noted that we reduce the dimension of the obtained representation from $401408$ ($56\times56\times128$) bins to $256$ bins using 2D average pooling layers followed by the flatten layer. More specifically, the network receives an n-channel input RGB-D image ($224\times224$ pixels) and by passing the image through the encoder part of the network, a features matrix of $56\times56\times128$ dimensions is generated. The obtained representation is then fed to two convolution layers having $31\times31, 64$ filters and $15\times15, 32$ filters with batch normalization and rectifier linear unit (ReLU) as an activation function, respectively. The second convolution layer has two identical branches as outputs: one is used for the growing dual-memory (GDM) learning, and the other is used for generating grasp maps. The features output from the second 2D convolution layer are further passed to three residual blocks, in which each block has two 2D convolution layers having $15\times15, 32$ filters with batch normalization. The residual block also uses ReLU as an activation function. 

For the GDM learning, we pass the output of the second convolution layer in encoder to the 2D average pooling layer followed by the flatten layer to obtain a feature vector. For the grasp synthesis, the output of encoder is passed to the decoder to generate a grasp map for the given object. It should be noted that the decoder architecture consists of two deconvolution (transpose-convolution) layers of size $15\times15, 32$ filters and $31\times31, 64$ filters with batch normalization and ReLU activation function, followed by the GR-ConvNet decoder. The output the GR-ConvNet decoder is three images of $224\times224$ dimensions to represent pixel-wise grasp quality, angle (cos + sin), and width. To train the network, we use smooth L1 loss function and RMSprop optimizer. The learning rate for RMSprop optimizer has been set to $0.001$ during training.

%The learning rate has been set to $0.001$ and the control parameters for exponential decay ($\beta_1$, and $\beta_2$) have been set to $0.9$, and $0.999$, respectively. 

%Using the obtained grasp map, the grasp point and locations are calculated by estimating the local peaks in the quality of the image predicted. The grasp rectangles are determined based on the estimated grasp points location (based on the number of grasps required) and the predicted angle and width images.

%In order to reduce the dimensions of we added two convolution layers of $256$ kernels of size $3\times3$ with stride $2$ and padding $1$ and the another convolution layers with $256$ kernels of size $1\times1$, at the end of the output from the residual block of the GR-ConvNet. 

The episodic memory of the dual-memory network receives the input feature vector, $x^i(t) \in \mathbb{R}^{256}$, to learn the similarities between feature vectors in an unsupervised fashion. The discrepancy between the sequential input and neural representation is minimized by creating new neurons or updating the existing neurons based on the activation threshold, habituation rate of the neurons, and other hyper-parameters (explained in section \ref{sec:4.2}). The learned weights from the episodic memory $W^{EM}_b$ (based on the total number of neurons) are then passed to the semantic memory to learn the task-specific knowledge. Thereby episodic memory results in the prediction of instance-level information and the semantic memory with the category level prediction. 

\subsection{Dual-memory recurrent self-organization network}
\label{sec:4.2}
Both the episodic and semantic memories use the Gamma-GWR \cite{parisi2017lifelong} network, which dynamically grows or shrinks based on the input data distribution. The neural network structure of the Gamma-GWR is recurrent, where the connection between the input data distribution and the best matching unit (BMUs) are determined based on the similarity measures (e.g. Euclidean distance). Neurons may also have more than one neighboring relation based on the similarity between the sensory information. Furthermore, the gamma memory in the Gamma-GWR holds the temporal relation of the neural activation trajectories during learning which gets dynamically changed relative to the input data distribution. This temporal memory is used during pseudo-rehearsal or intrinsic replay to alleviate the catastrophic forgetting in the incremental learning task. The networks are initialized with two neurons and it dynamically grows by holding the spatio-temporal relation while iterating over the input data samples. Each neurons A, in the network consist of weight vectors $w_j \in \mathbb{R}^n$ and K context descriptors $C_{k,j} \in \mathbb{R}^n$. For each given input $x^i(t) \in \mathbb{R}^n$ the Best Matching Unit, BMU ($b$), is calculated based on the Manhattan metrics (given in equation \ref{fromula-1} - \ref{formula-3}). It should be noted that Parisi et al., \cite{GDM/10.3389/fnbot.2018.00078} used euclidean distance as a similarity measure, while we analysed the performance of the prediction using different distance measure (Euclidean distance, Squared Euclidean distance, Manhattan distance, Minkowski distance with power of 3, Mahalanobis distance, and cosine similarity measure), based on experimental results we found that Manhattan distance metric suits best for our input data distribution. In particular, we calculate the dissimilarity and the best matching unit as follows:
\begin{equation}
    d_j = \alpha_0|x_i(t) - w_j| + \sum_{k=1}^K \alpha_k|C_k(t) - c_{j,k}| 
    \label{fromula-1}
\end{equation}
\begin{equation}
    C_k(t) = \beta . w_b^{t-1} + (1-\beta).c_{b,k-1}^{t-1}
    \label{formula-2}
\end{equation}
\begin{equation}
    b = arg \min_{j \in A}(d_j)
    \label{formula-3}
\end{equation}
where $\alpha$ and $\beta$ are two constants that regulate the temporal context influences, $w_b^{t-1}$ is the weight vector of the BMU at time $t-1$, $C_k \in \mathbb{R}^n$ is the global context descriptors with $C_k(t_0) = 0$, and $c_{b,k-1}^{t-1}$, the context descriptor of the BMU and $k-1$ descriptor at $t-1$. The neurons in each network are either created, or existing neurons are updated based on the activity of the neuron, $a(t)$ and its habituation counter $h_j$, regulated by the insertion threshold ($a_T$) and habituation threshold $h_T$. The activity of neuron $a(t)$ is determined based on the distance relationship between the input and BMU~($b$) which is computed as follows:  
\begin{equation}
    a(t) = \exp(-d_b)
    \label{formula-4}
\end{equation}
when the neuron with its respective BMU predicts the input sequence correctly, results in the highest activation value of $1$. The habituation counter ($h_j \in [0,1]$) expresses the frequency of the neuron firing in the training process. The habituation values of the BMU~($b$) and it neighbour (n) decreases as the frequency of the neuron firing increase. Compared to other conventional unsupervised learning algorithms, where the winner takes all the credits, in GWR not only the winner but also its associated neighbouring neurons are updated. The habituation rule~\cite{MARSLAND20021041} for a neuron $i$ is given by:
\begin{equation}
    \Delta h_i = \tau_i.\kappa.(1-h_i) - \tau_i 
    \label{formula-5}
\end{equation}
\noindent where $\tau_i$ and $\kappa$ are two constants that control the monotonically decreasing behavior of the habituation counter, the habituation counter of BMU ($b$) decreases faster than neighbouring neurons ($n$). The weight vectors and the context descriptors are get updated where ever the new neurons are created and existing nodes are updated, when new neurons are created its weights are computed as the average weights of BMU and input. The weight and context descriptor update for the neuron i is given as:
\begin{equation}
    \Delta w_i = \epsilon_i.h_i(x(t)-w_i)
    \label{formula-6}
\end{equation}
\begin{equation}
    \Delta c_{i,k} = \epsilon_i.h_i(C_k(t)-c_{i,k})
    \label{formula-7}
\end{equation}
\noindent where $\epsilon_i$ is a learning rate, the learning rate of BMU ($b$) will be higher than the neuron ($n$). The connection between the two neurons (BMU and second BMU) is created when two neurons fire together. Each neuron that exists in the network has a certain age, when those ages reached a certain threshold it will be removed from the network. The ages between the first BMU and second BMU rest to zero, whereas other neighbouring ages are increased by the value of $1$. At the end of training epochs, the neurons with an age higher than the threshold and the neurons which do not have neighbours are removed from the network. \\ \\ 
\noindent \textbf{Episodic memory:}
In episodic memory (G-EM), the neuron's growth is unsupervised. Based on the insertion threshold, $a_T$, and the habituation threshold, $h_T$, the network learns a fine-grained representation of the input data since new neurons will be created when the activity of the BMU falls below the insertion threshold. The temporal connection of the neural activation trajectories are learned during episodic memory training by sequence selective synaptic links~\cite{GDM/10.3389/fnbot.2018.00078}. When two neurons are activated continuously their temporal synaptic link, $p(i,j)$, are increased by $\Delta p_{(i,j)} = 1$. For each neuron $i \in A$, the next neuron $v$ of a prototype vector can be retrieved by selecting $v=\operatorname*{argmax}~p(i,j)$, where $i$ and $j$ represent the neurons at time $t-1$ and $t$. The temporal representation learned in episodic memory are used during memory replay in the absence of external sensory input. During the learning phase, the neurons in G-EM learn the instance-level label ($l^I$) representation. The associative matrix, $H(j,l^I,l^C)$ \cite{GDM/10.3389/fnbot.2018.00078} stores the label information for each neuron $j$. Therefore, using the associative label representation, the unsupervised Gamma-GWR can be used for classification tasks without the pre-determined number of labels. \\ \\
\noindent \textbf{Semantic memory:} 
Similar to episodic memory, semantic memory (G-SM) is also associated with the Gamma-GWR network. Instead of receiving the direct sensory information as input, G-SM received the episodic weights $w_b^{EM}$ as input, based on the number of neurons in the G-EM and the associative matrix containing the label information, $l^C$. Unlike G-EM, G-SM creates new neurons only when the labels predicted by the BMU in G-SM are miss-classified with the ground truth labels. Since G-SM uses category-level signals to regulate the network growth, the same neurons may be activated for the different instances of the object which belongs to the same category. In this way, G-SM develops a compact representation from the episodic experience~\cite{GDM/10.3389/fnbot.2018.00078}. \\ \\
\noindent \textbf{Memory replay:} We use memory reply techniques to handle catastrophic forgetting by reactivating the past neural activation pseudo-patterns. The pseudo-patterns are learned during the G-EM training and are represented as temporally ordered neural activation trajectories. These trajectories is then replayed using pseudo-rehearsal or intrinsic replay \cite{robins1995catastrophic}.  Recursively reactivate neural activity trajectories (RNATs) \cite{GDM/10.3389/fnbot.2018.00078} using temporal synapses in G-EM, the neural trajectories are computed to each neuron in the episodic memory for the fixed temporal window and replayed back to the G-EM and G-SM after each learning episode. RNATs computation doesn't explicitly need storing the temporal relation and the labels of the previously seen samples to remember the past knowledge since it generates the sequence-selective prototype sequence during each learning iteration which will be periodically replayed back to G-EM and G-SM networks.
\\ \\
\noindent \textbf{Controlled neuron connections removal}: 
\label{sec:controlled_removal}
In the work by Parisi et al., \cite{GDM/10.3389/fnbot.2018.00078}, at the end of each epoch the connection between the neighboring neurons are removed when the age of a particular node reached its maximum. Therefore, at the end of the training, the neurons without neighbours are also removed.  When neurons are removed from the network the information gained about the input data is also completely removed, regardless of that neurons knowledge level. Such types of removal is not vital. For example, consider a scenario that at the beginning of learning process a neuron learns useful information for a particular object category, and the neuron has only one neighbour connection and is not triggered further. The age of the neuron increases as the training process continues, and consequently, at the end of training the neuron will be removed due to the high age despite its knowledge about the input data. We address this problem by using the regulated neuron connection removal. The connection between the neighboring neurons is removed when its habituation value (which holds the knowledge level of the neuron) is greater than the certain threshold value. If the neuron habituation value is greater than a threshold value and its age is higher than a maximum age, the ages and connection of its associated neighbours are set to $0$. If the habituation value of the neuron is less than a threshold and its age is greater than a threshold, the connection between its neighbours is retained only their ages are rest to $0$. In this way, we can retain the neurons which have a better understanding of the input data. From the experimental observations, we observed that the controlled neuron removal helps to significantly improve the performance in instance- and category-level accuracy (see section~\ref{controlled_connections_removal}).  
%%%%%%%%%%%%%%%%%%%%%%%%%%%%%%%%%%%%%%%%%%%%%%%%%%%%%%%%
\section{Experimental Results}
\label{experiments_results}

We evaluated the performance of the proposed system in three rounds of experiments: (\textit{i}) the performance of the autoencoder model is evaluated using the Cornell dataset~\cite{5980145}, (\textit{ii}) we then assessed the performance of the GDM learning on batch, incremental, and continuous object recognition scenarios, and (\textit{iii}) finally, the overall system (see Fig. \ref{fig:1}) is tested on the simulated robot environment\footnote{Code and supplementary material are available at: \url{https://github.com/krishkribo/3D_GDM-RSON}.}. 

%\cred{ --- add a brief intro here --- }

% short over all description
\subsection{Dataset Generation}
\begin{wrapfigure}{r}{0.4\textwidth}
    \centering
    \includegraphics[width=\linewidth]{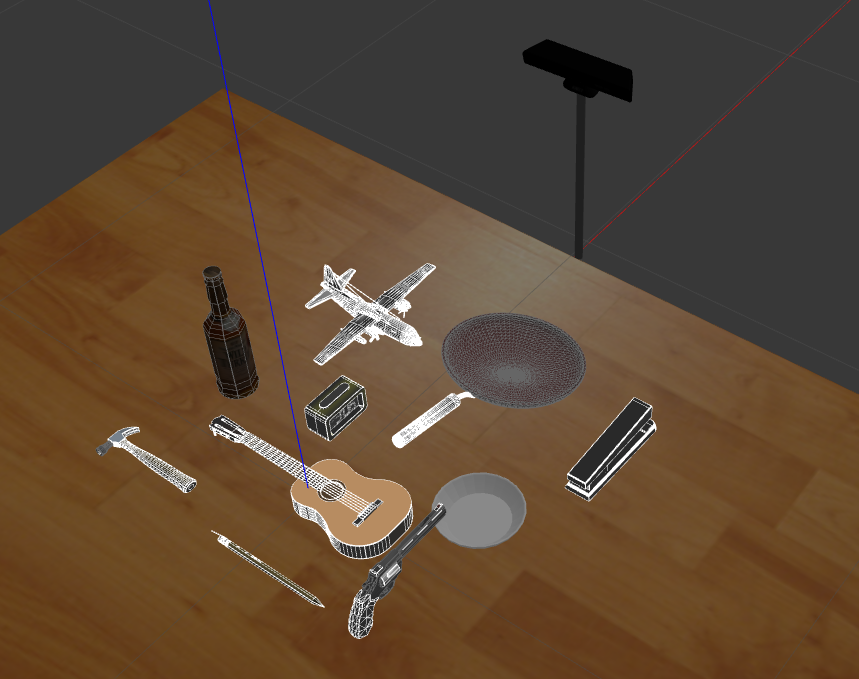}
    \caption{Dataset generation environment setup, Kinect camera with sample objects from $10$ categories.}
    \label{fig:4}
    \vspace{-1mm}
\end{wrapfigure}
\begin{figure} [!b]
    \centering
    \includegraphics[width=0.9\linewidth]{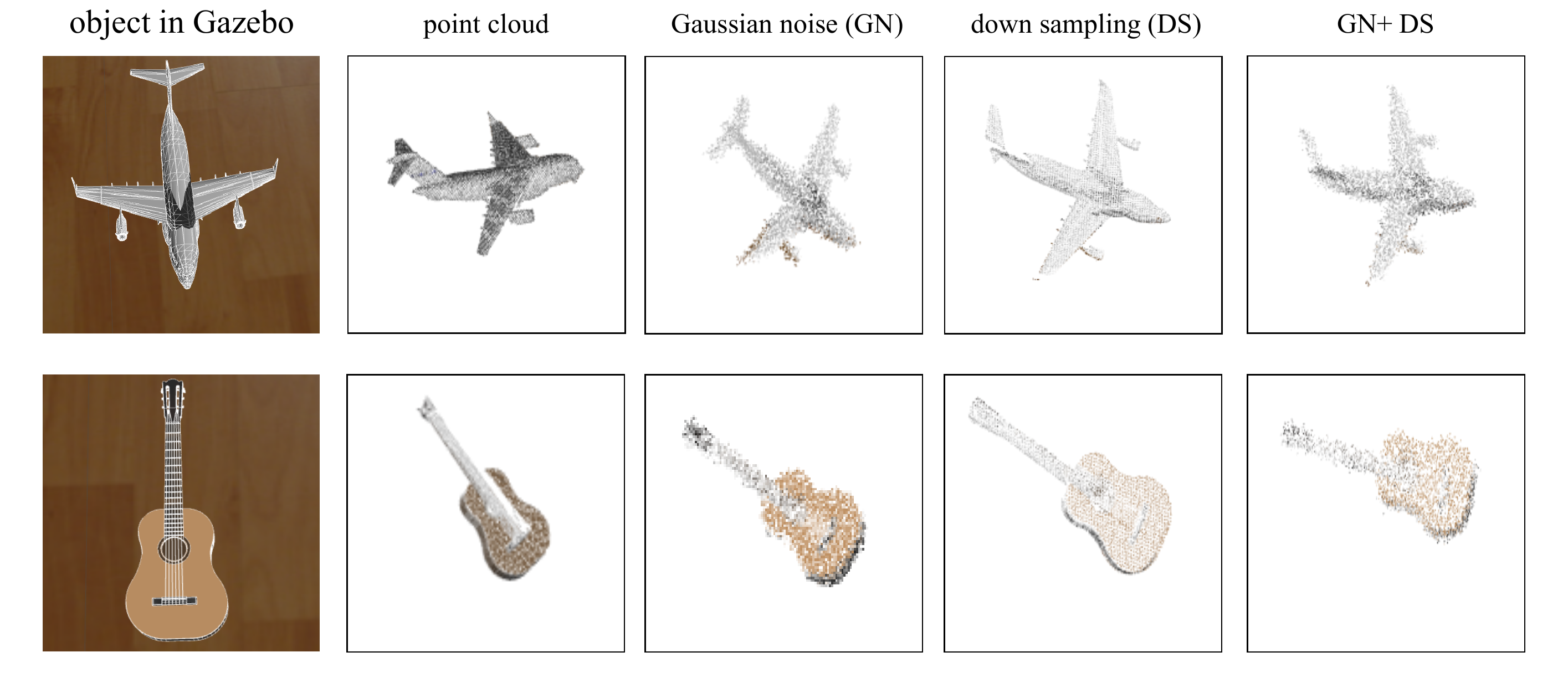}
    \caption{Data augmentation:  airplane (\textit{top-row}), and guitar (\textit{bottom-row}) in the Gazebo environment followed by their point clouds. For augmenting the data, we have applied: (\textit{i}) Gaussian noise, (\textit{ii}) down-sampling, and (\textit{iii}) Gaussian noise plus down-sampling to the point cloud of the object.}
    \label{fig:5}
\end{figure}

We generated a large-scaled synthetic sequential point cloud dataset for continuous object recognition task. 
%Our setup has been developed over Robot Operating System (ROS) and Gazebo simulation environment.
To generate the dataset, we used objects from the ShapeNet dataset~\cite{chang2015shapenet} and Gazebo repository. The dataset comprises of $50$ objects instances from $10$ categories in the form of sequential point cloud samples. Figure~\ref{fig:4} shows our environmental setup in Gazebo environment, and $10$ sample object categories. In this work, we apply 15 augmentation techniques, such as a change in translation and position of the object, adding Gaussian noise with a different standard deviation, down-sampling with different resolutions, and adding occlusions, to enhance the size of the dataset\footnote{The complete dataset generation pipeline, the techniques and the parameters used to generate the dataset are explained in the supplementary material.}. Two sets of augmented point clouds for Airplane and Guitar objects, generated using the different techniques, are shown in Fig.~\ref{fig:5}. By visualizing the obtained augmented objects, we can observe that the overall structure of the object is retained. At the end of dataset generation processes, we created $15$ collections of data, where each collection contains sequential data for all 10 object categories, i.e., generated at the rate of $2.5$ fps, that only one augmentation technique applied to the objects. In total, we generated $75000$ samples, which required $1.7$ GB data storage. This dataset is used to train and evaluate the GDM model in our work. In particular, from the 15 generated collections, we used $12$ collections for training, and three collections are used for testing. 
%%%%%%%%%%%%%%%%%%%%

\subsection{Grasp Network}

The proposed model is developed over PyTorch library. We trained the model on the NVIDIA RTX-2070 Max-Q GPU, and the training procedure took around five hours for $50$ epochs.  

\begin{figure}[th]
    \centering
    \includegraphics[width=\linewidth]{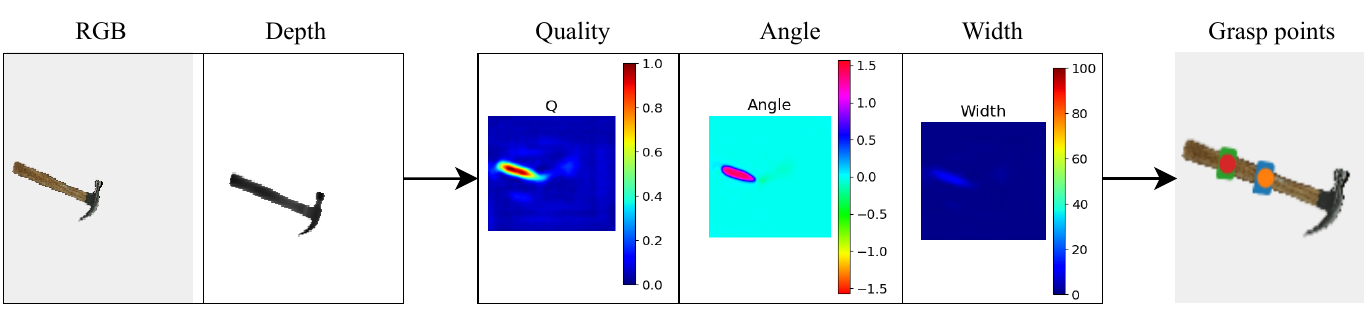}
    \caption{The middle three images show the pixel-wise outputs of the proposed autoencoder for a hammer object. The top-two grasp configurations for the hammer object are shown in the right-most image. The colored circles indicates the grasp point which is located at the rectangle center. The blue rectangle represents the grasp's orientation and width for the first grasp point while the green one shows for the grasp configuration for the second best point.}
    \label{fig:autoencoder_res}
\end{figure}

The network receives a 4-channel $224\times224$ RGB-D images as an input, which are sampled from the input point cloud data. The network is trained using Cornell grasping dataset~\cite{5980145}. We augmented the Cornell dataset with random rotation and random zoom to increase the sample size, and used $90\%$ of the data for training and $10\%$ for testing.
Root Mean Square propagation (RMSprop) is used  for training the model for $50$ epochs with the batch size of $8$. To handle the exploding gradients, a smooth $L1$ loss function is used. The model is then evaluated based on the Intersection over Union (IoU) score. A predicted grasp configuration is considered as correct when the overlap of the ground-truth grasp rectangle and predicted grasp rectangle is more than $25\%$, and the grasp orientation offset between ground-truth and the predicted grasp rectangle is less than $30$ degree~\cite{5980145}. Figure \ref{fig:autoencoder_res} shows the pixel-wise outputs of the network in terms of grasp quality, grasp angle, and grasp width for a given hammer object. The two best grasp configurations are highlighted in the right-most image.

\begin{wraptable}{r}{0.4\textwidth}
\vspace{-4mm}
\caption{Result of autoencoder network on the Cornell dataset~\cite{5980145}}
\resizebox{\linewidth}{!}{%
\begin{tabular}{@{}lll@{}}
\toprule
\textbf{Approach}  & \textbf{Input data} & \textbf{IoU (\%)} \\ \midrule
GG-CNN~\cite{morrison2018closing}                 & depth image & 73.00 \\
GG-CNN2~\cite{doi:10.1177/0278364919859066}$^*$   & depth image & 75.20 \\
GraspNet~\cite{ijcai2018-677}                     & RGB-D       & 90.20 \\
GR-ConvNet~\cite{kumra2021antipodal}$^*$          & RGB-D       & 96.00 \\
Our approach                                      & RGB-D       & \textbf{97.75} \\ \bottomrule
\textit{$*$ retrained.}
{\ul }   
\end{tabular}}
\vspace{-2mm}
\label{table:autoencoder_res}
\end{wraptable}

Table \ref{table:autoencoder_res} shows the comparison result of our approach with four state-of-the-art grasping methods: GG-CNN~\cite{morrison2018closing}, GG-CNN2~\cite{doi:10.1177/0278364919859066}, GraspNet~\cite{ijcai2018-677}, and GR-ConvNet~\cite{kumra2021antipodal} (retrained). By comparing all the obtained results, it is visible that our approach obtained the best overall IoU accuracy of $97.75 \%$. In particular, the proposed method worked $24.75$, $22.55$, $7.55$ percentage point (p.p) better than GG-CNN~\cite{morrison2018closing} , GG-CNN2~\cite{doi:10.1177/0278364919859066}, and GraspNet~\cite{ijcai2018-677} respectively. Since the difference in IoU metric between our approach and the GR-ConvNet model was small, we retained the GR-ConvNet model to provide a fair comparison. Based on our experiments, we noticed that our approach shows $1.75 $ p.p improvement in performance compared to GR-ConvNet model. This improvement can be correlated to the size of the network, as our model has $\sim 2.3\operatorname{M}$ (million) trainable parameters whereas the GR-ConvNet model has $\sim 1.9 \operatorname{M}$ parameters. More specifically, our custom network has around $400 \operatorname{K}$ more trainable parameters than the GR-ConvNet model. As shown in Fig.~\ref{fig:3}, the learned latent representation of the network is then used as an input to the episodic memory.

\subsection{GDM Learning}
We performed series of experiments to evaluate the performance of the GDM networks in batch and incremental learning scenarios. The model is trained and tested with our generated sequential dataset for instance- and category-level prediction. We also tested the network learning based on the temporal context and intrinsic memory replay \cite{GDM/10.3389/fnbot.2018.00078}. Furthermore, a set of experiments has been carried out to evaluate the continual learning task in the incremental learning scenario. The evaluation scenario includes testing new instances (NI) of already known category, testing never-seen-before object categories (NC) during the training processes, and testing new instance and new category (NIC) which are progressively introduced during learning.

\subsubsection{Batch Learning}
In batch learning, the dataset consists of all the objects from all the collections that are used for training. The performance of the model is evaluated on the test dataset on an instance and category-level object recognition.

We experimented the batch learning under three different conditions GDM with TC, GDM without TC during testing (by setting context descriptor $K=0$), and GDM without TC. In our experiments, the number of context descriptors is set to two for both G-EM and G-SM $(K^{EM},~K^{SM})$ networks. Based on the selection of context descriptors, the memories in GDM activates for the temporal window of number of context descriptors $[K + 1]$. In our experiments, the G-EM has $2$ context descriptors, hence it activates for the temporal window of $3$ input frames $[(K^{EM} = 2) + 1]$. Since G-SM receives input as neurons from G-EM, it codes for the temporal window of $5$ frames $[(K^{EM} = 2) + (K^{SM} = 2) + 1]$.

\begin{wraptable}{r}{0.5\textwidth}
\vspace{-4mm}
\caption{Classification accuracy of batch learning on category-level for different approaches.}
\resizebox{\linewidth}{!}{%
\begin{tabular}{@{}lll@{}}
\toprule
\textbf{Approach}  & \textbf{Accuracy (\%)} & \textbf{Accuracy (\%)} \\
& \textbf{(Training)} & \textbf{(Testing)}
\\ \midrule
GDM with TC     & {$93.35\pm 0.09$} & {$88.53\pm 0.04$} \\
GDM* without TC$^1$  & {$92.98 \pm 0.09$} & {$87.23 \pm 0.05$} \\
GDM without TC & {$90.52 \pm 0.07$} & {$87.35 \pm 0.06$} \\ \bottomrule
{\ul }   
\end{tabular}}
\footnotesize{\textit{$^1$ GDM* model trained without TC during testing.}}
\label{table:batch}
\end{wraptable}
The number of context descriptors is determined based on different experimental observations. For batch learning, we synthesized $120$ samples from each object category to train the GDM model (i.e., $3000$ samples in total). The obtained results of the batch learning on category-level are reported in Table~\ref{table:batch}. It should be noted, the results are obtained after 35 epochs, which are averaged across five learning trials by randomly shuffling the bathes from different collections.

GDM with temporal context gives the best accuracy compared to the other two approaches. We obtained average accuracy of $86.97 \%$ (instance-level) and $93.35 \%$ (category-level). When tested on the never-seen-before objects, we obtained an average accuracy of $88.53 \%$. When compared with the results of GDM training without temporal context, the instance-level accuracy shows an improvement over $5.1 \%$ and the category-level accuracy shows $2.83 \%$ improvement. Based on these results, it is clear that learning the temporal relation of the input plays important role in increasing the performance of the model. We also observed that the performance of the model slightly dropped in both training and testing phases if the model does not use temporal context.
%during the learning, which shows slight improvement ($2.95 \%$) while validating, but gives more or less equal results while testing ($79.82 \%$ - without TC test, and $80.07 \%$ - without TC).

\begin{figure}[!t]
\hspace{-5mm}
    \begin{tabular}{cccc}
        \includegraphics[width=0.25\textwidth]{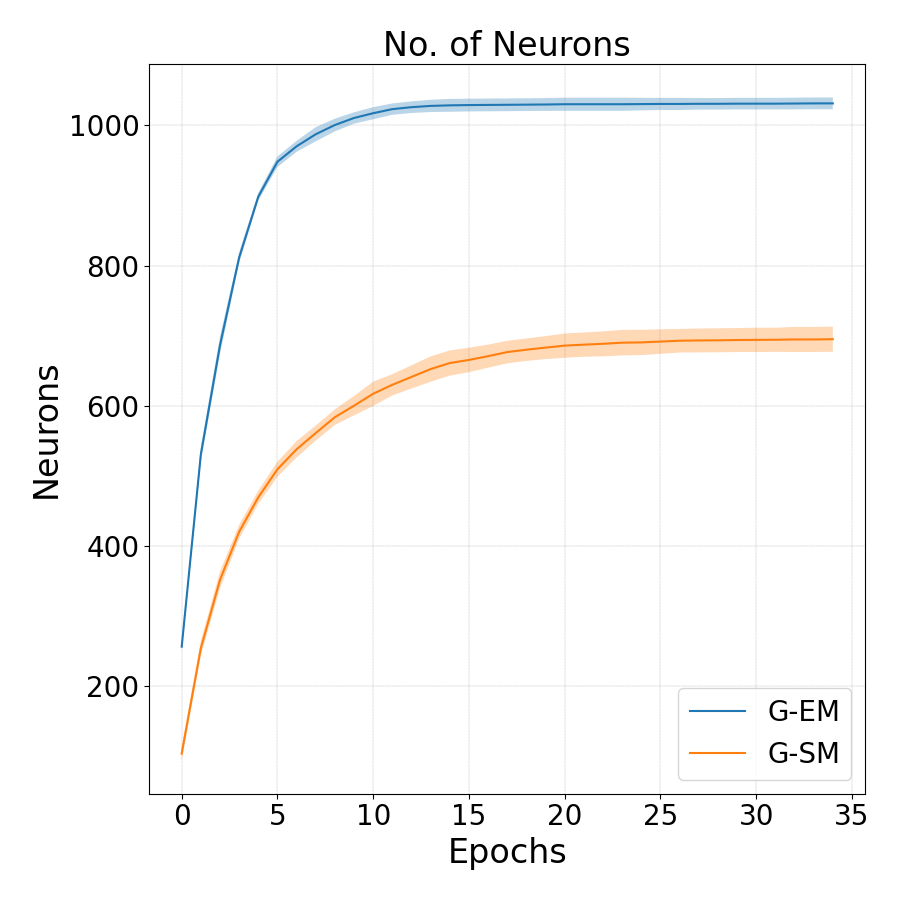} &\hspace{-5mm}
        \includegraphics[width=0.25\textwidth]{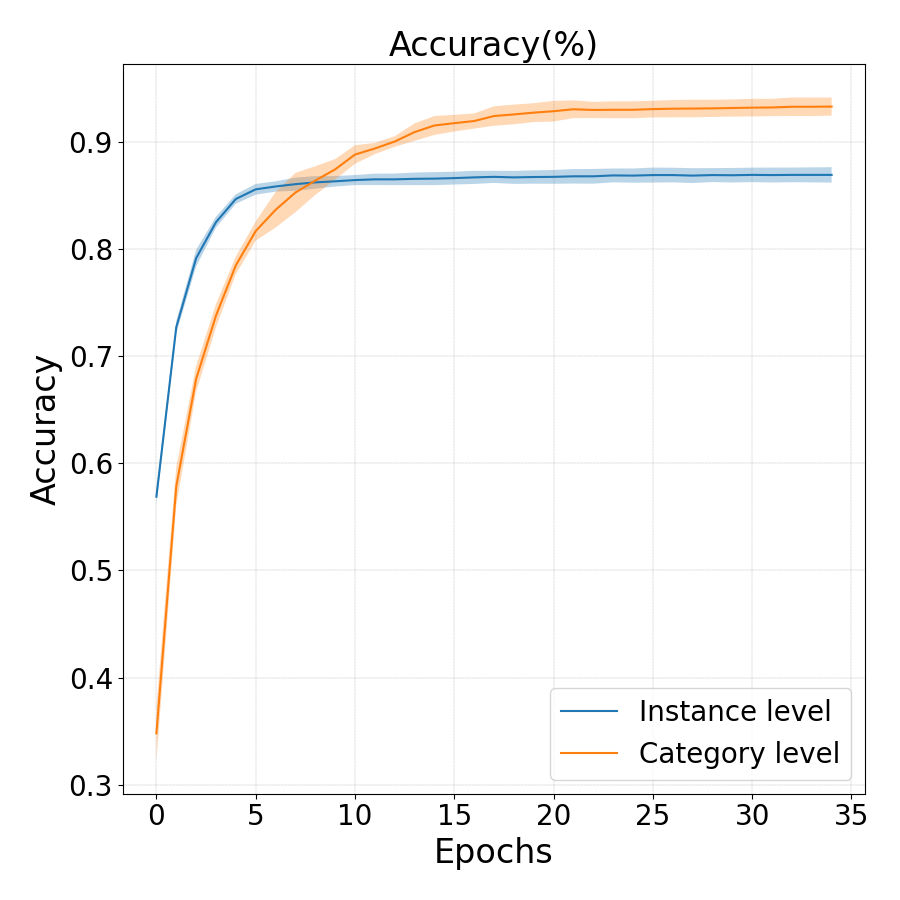}  & \hspace{-5mm}
        \includegraphics[width=0.25\textwidth]{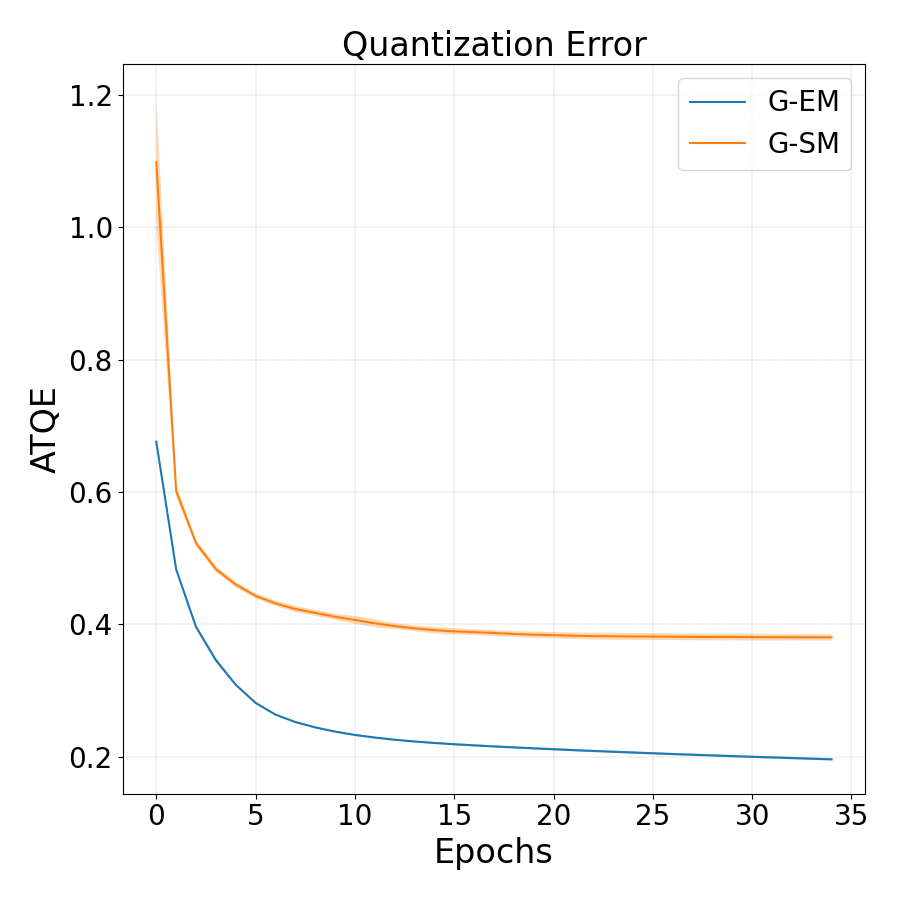} &\hspace{-5mm}
        \includegraphics[width=0.25\textwidth]{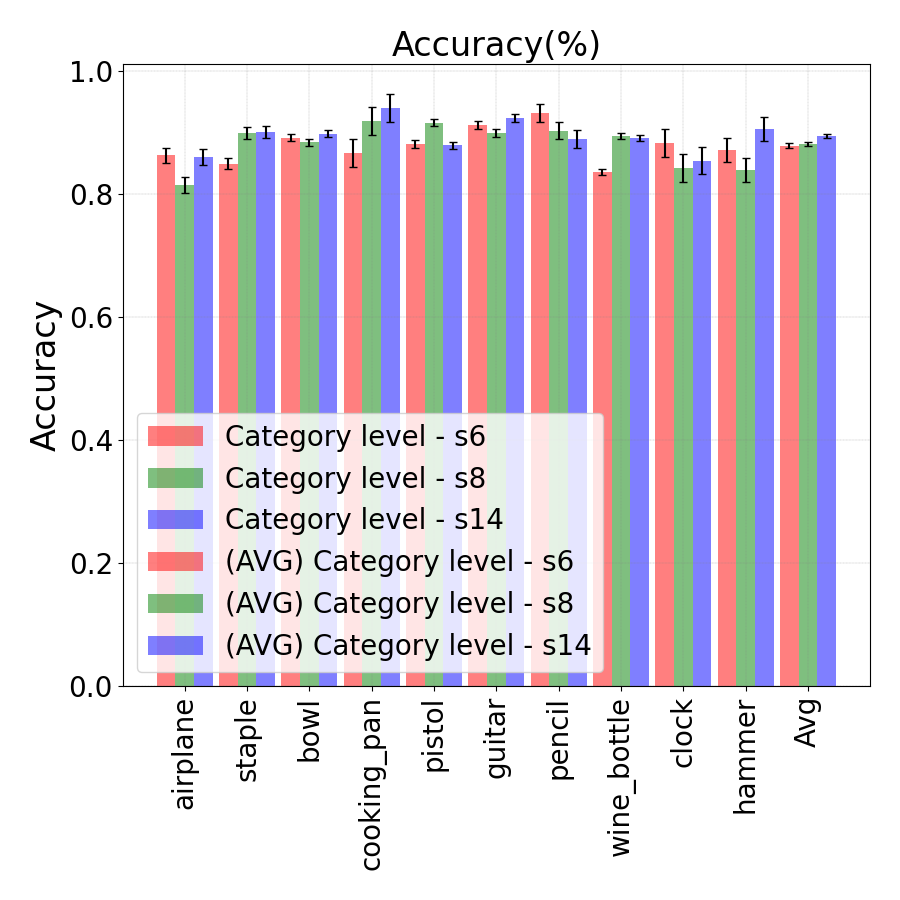}\\
        (a)& (b)& (c)& (d)
    \end{tabular}
    \caption{Batch learning: (a) Number of neurons in episodic and semantic memory as a function of number of epochs, (b) Accuracy of G-EM and G-SM on instance-and category-level, (c) Quantization error, and (d) accuracy on the test dataset. Results are averaged over five trials, and the shaded areas show the standard deviation.}
    \label{fig:8}
\end{figure}

From the results, we can conclude that model not only performs better at training data but it also shows good generalization results on the test data. Figure~\ref{fig:8} shows the number of neurons, accuracy, average quantization error, and category-level accuracy on the test collections through $35$ epochs and averaged across five learning trials. Fig.~\ref{fig:8}(a) shows that the growth of the neurons is stabilized after $15$ epochs for both episodic and semantic memory. This indicates that the neurons are habituated for the given input data. It can be observed that the number of neurons in episodic memory is higher than semantic memory. 
This is expected since the G-SM network grows only when the predicted BMU label is misclassified with the input data label, whereas G-EM growth is unregulated, and it can grow when a new data is given. As the number of neurons increased and habituated to the input data, we can see that the average quantization error in Fig.~\ref{fig:8}(c) is significantly reduced and the accuracy over the epochs also getting stabilized (Fig.~ \ref{fig:8}b). From the result of individual object category accuracy on the test collection (see Fig.~\ref{fig:8}d), it can be seen that most of the objects are classified with a higher accuracy. 
%Except for few objects like pencil, fork, and USB where the prediction performance is comparatively less than other object categories. This may be due to the existence of identical object categories like spoon, pen, staple, and eraser. Even though we observed performance drop for those objects, it should be noted that the GDM model is able to identify and distinguish the objects with similar shapes and sparse representations. 

\subsubsection{Incremental Learning}

In this round of experiments, the training samples are progressively available in the form of mini-batches. Each mini-batch contains samples from all training collections based on the different object instances and categories. In this experiment, we defined the number of epochs equal to the number of categories, and in each epoch we use objects from a certain category to train the model. This way, instances of previously learned categories are not shown again while learning new categories.

The behaviour of the network in alleviating the catastrophic forgetting is evaluated by using the recursive reactivate neural activation trajectories (RNAT's) and intrinsic memory replay. We then assessed and compared the models trained with and without memory replay.

After introducing each object category, the accuracy of the model is evaluated using never-seen-before-samples from all known categories to check if the model has learned all categories accurately. For example, if the network is training on the third object category (i.e. at third epoch), the performance of the model is evaluated using data samples from all learned categories( i.e., test samples are the first, the second, and the third categories). This way, the performance of the model can be estimated to check whether the model learned new object categories without forgetting the previously learned ones. 

%\cred{what happen if the the model can not recognize objects correctly after introducing a new categories?} {We observe a drop in accuracy (on that particular category), since the model is unsupervised it continually moves to learning next object category. In figure \ref{fig:9} (b) the small low hill in category level accuracies is the indication of that the model could not recognize all the samples representing that particular object category. The reason we discussed for this accuracy drop is the existence of low feature and identical object categories like spoon, pencil, fork, eraser, staple, and spoon.}

\begin{wraptable}{r}{0.5\textwidth}
%\centering
\caption{Classification performance of incremental learning at category-level during training and testing.}
\resizebox{\linewidth}{!}{%
\begin{tabular}{@{}lll@{}}
\toprule
\textbf{Approach}  & \textbf{Accuracy (\%)} & \textbf{Accuracy (\%)} \\
& \textbf{(Training)} & \textbf{(Testing)}\\ \midrule
GDM with replay     & {$73.80 \pm 0.11$} & {$70.01 \pm 0.12$} \\
GDM without replay  & {$56.75 \pm 0.19$} & {$58.70 \pm 0.26$} \\ \bottomrule
{\ul }       
\end{tabular}
}
\label{table:inc}
\end{wraptable}
We trained the network for $10$ epochs, as the dataset has $10$ categories, and repeated the experiments for five times to have statistical significance of the results. The obtained results of incremental learning with and without memory replay are summarized in Table~\ref{table:inc}. As reported in table \ref{table:inc}, GDM with intrinsic memory replay (MR) obtained the best overall average accuracy, which is $57.27 \%$ accuracy at the instance-level, $73.80 \%$ at the category-level on the validation set, and $70.01 \%$ accuracy on the test samples. By comparing the results of the GDM model trained with and without memory replay, we observed $2.74 \%$ reduction in instance-level accuracy while the category-level accuracy increased by $17.05 \%$. This increase in category-level is also observed when both models tested on never-seen-before samples. IN particular, we observed that the model with MR gives $11.31 \%$ accuracy improvement over the model without MR. 

\begin{figure}[!t]
\hspace{-5mm}
    \begin{tabular}{cccc}
        \includegraphics[width=0.25\textwidth]{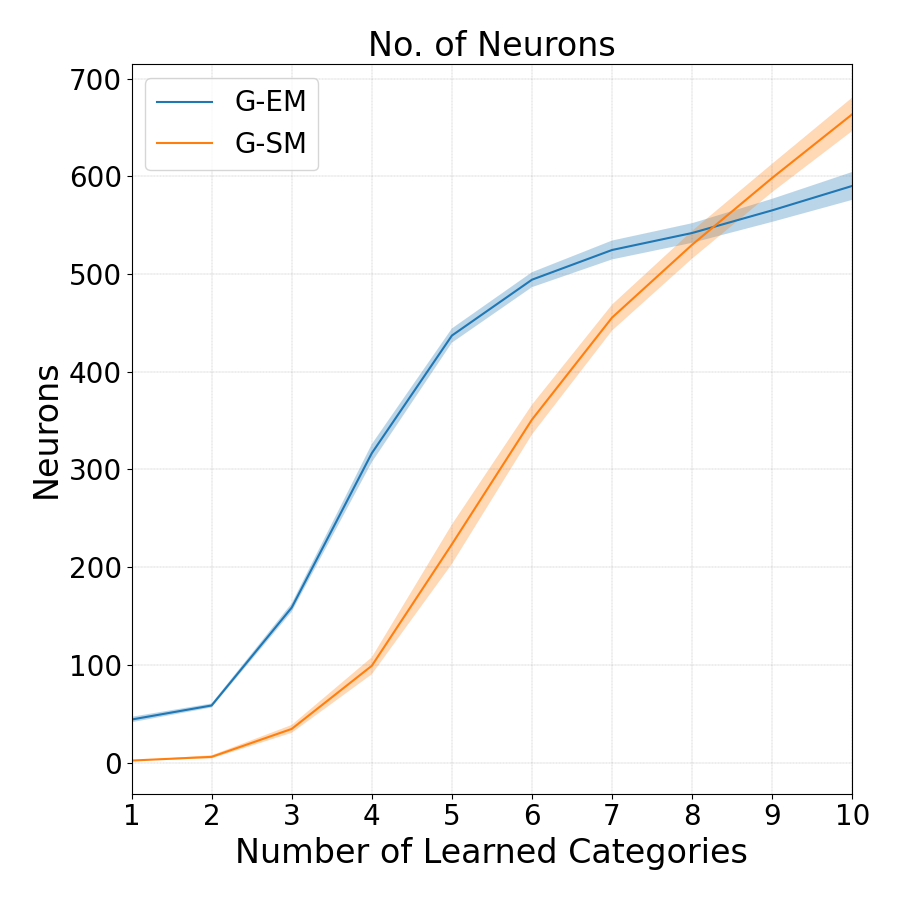} &\hspace{-5mm}
        \includegraphics[width=0.25\textwidth]{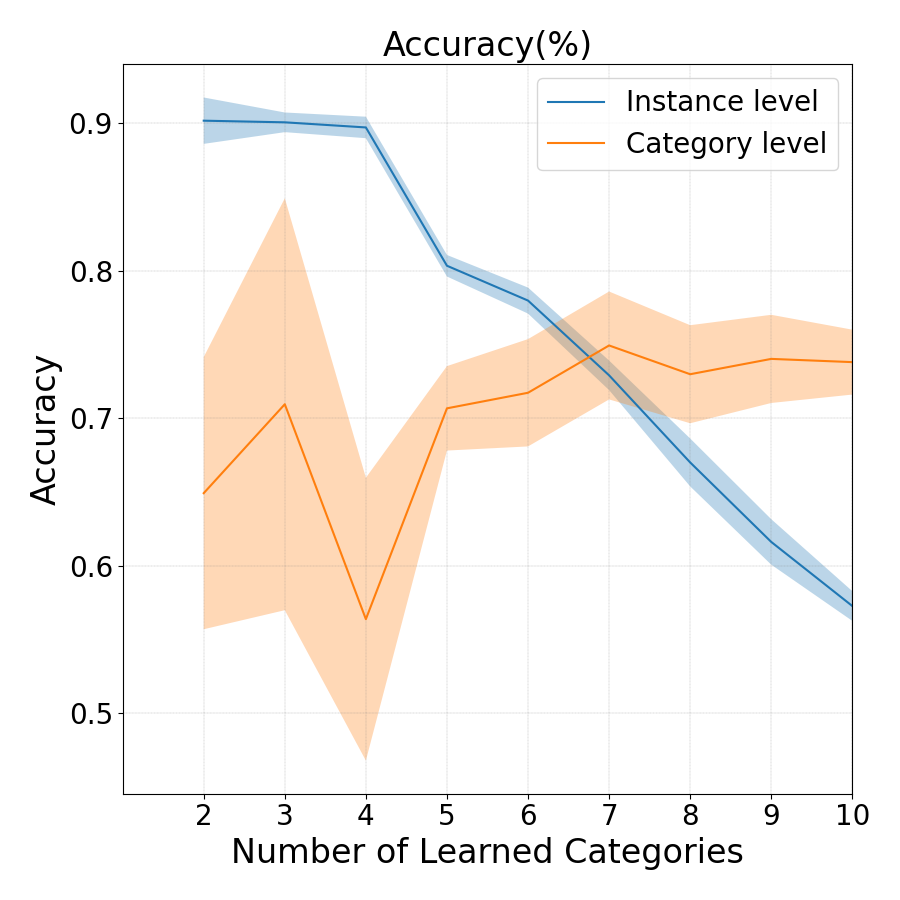}  & \hspace{-5mm}
        \includegraphics[width=0.25\textwidth]{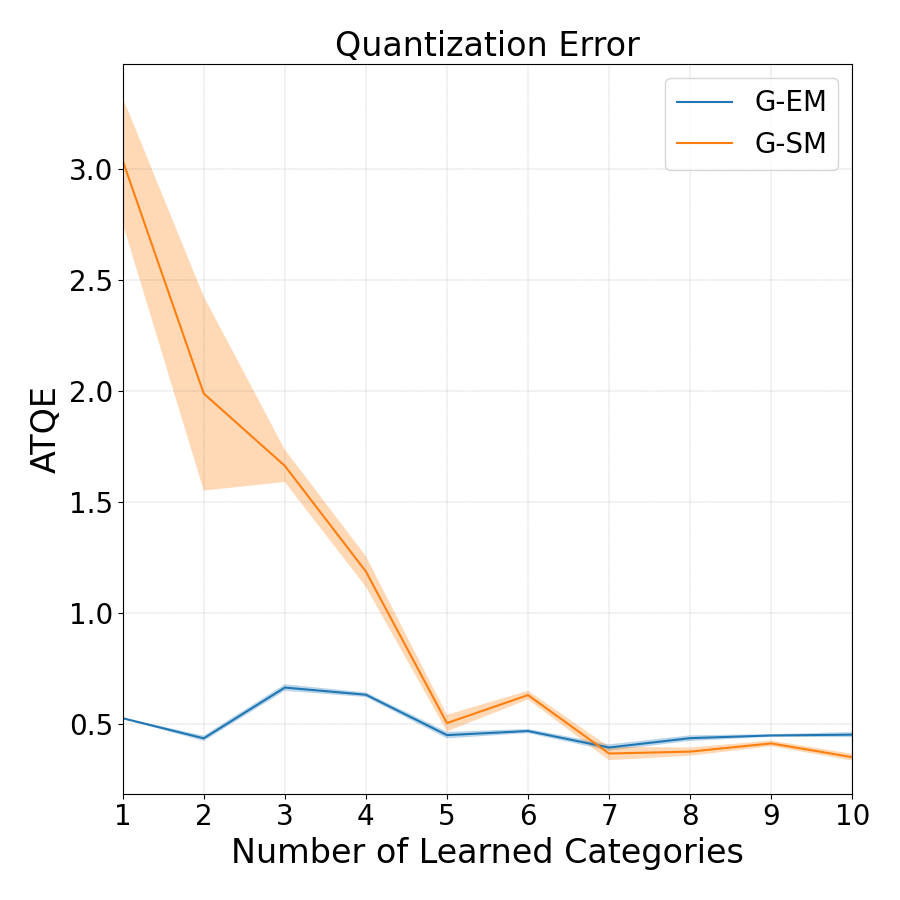} &\hspace{-5mm}
        \includegraphics[width=0.25\textwidth]{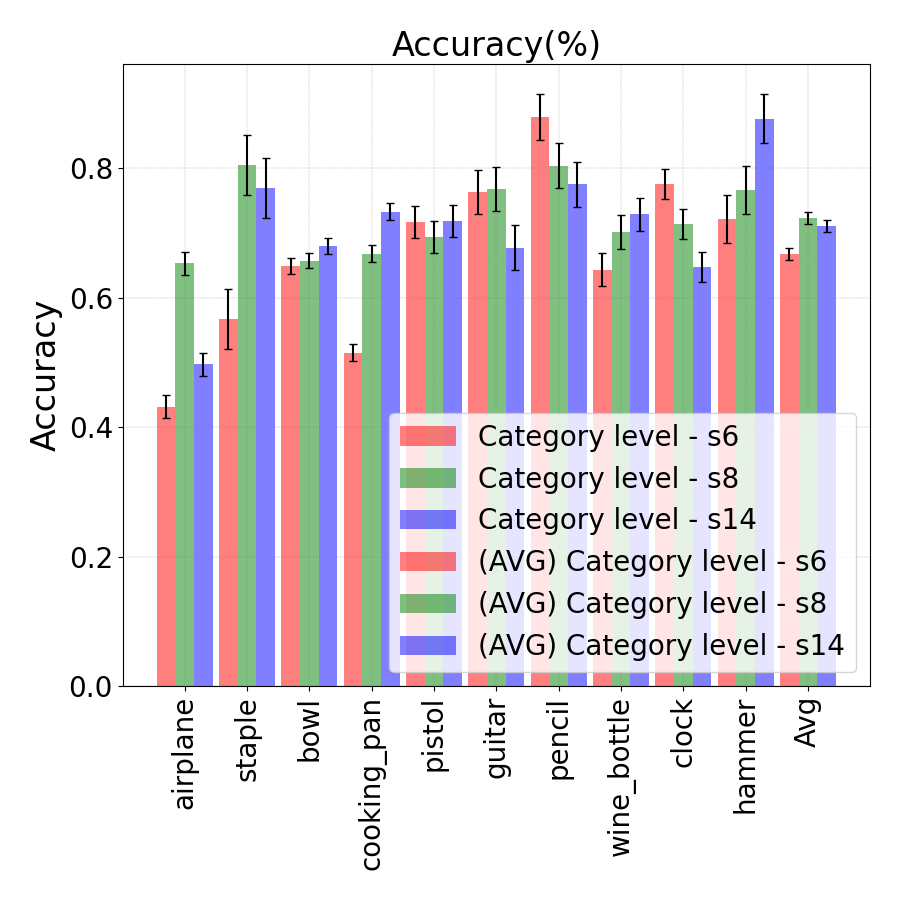}\\
        (a)& (b)& (c)& (d)
    \end{tabular}
    \caption{Incremental learning results: (a) Number of neurons in episodic and semantic memory, (b) Accuracy of G-EM and G-SM on instance and category level, (c) Quantization error, and (d) Accuracy on the test dataset. Results are averaged for five learning trials, the shaded area shows the standard deviation.}
    \label{fig:9}
\end{figure}

\begin{figure}[!t]
\hspace{-5mm}
    \begin{tabular}{cccc}
        \includegraphics[width=0.25\textwidth]{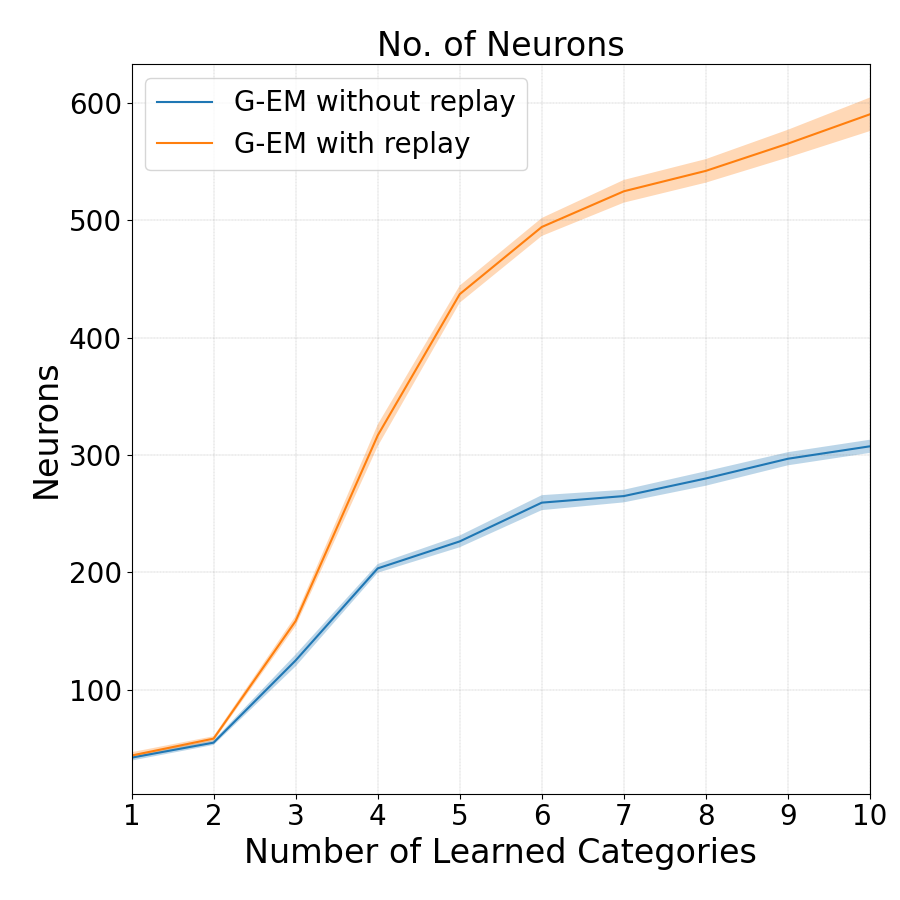} &\hspace{-5mm}
        \includegraphics[width=0.25\textwidth]{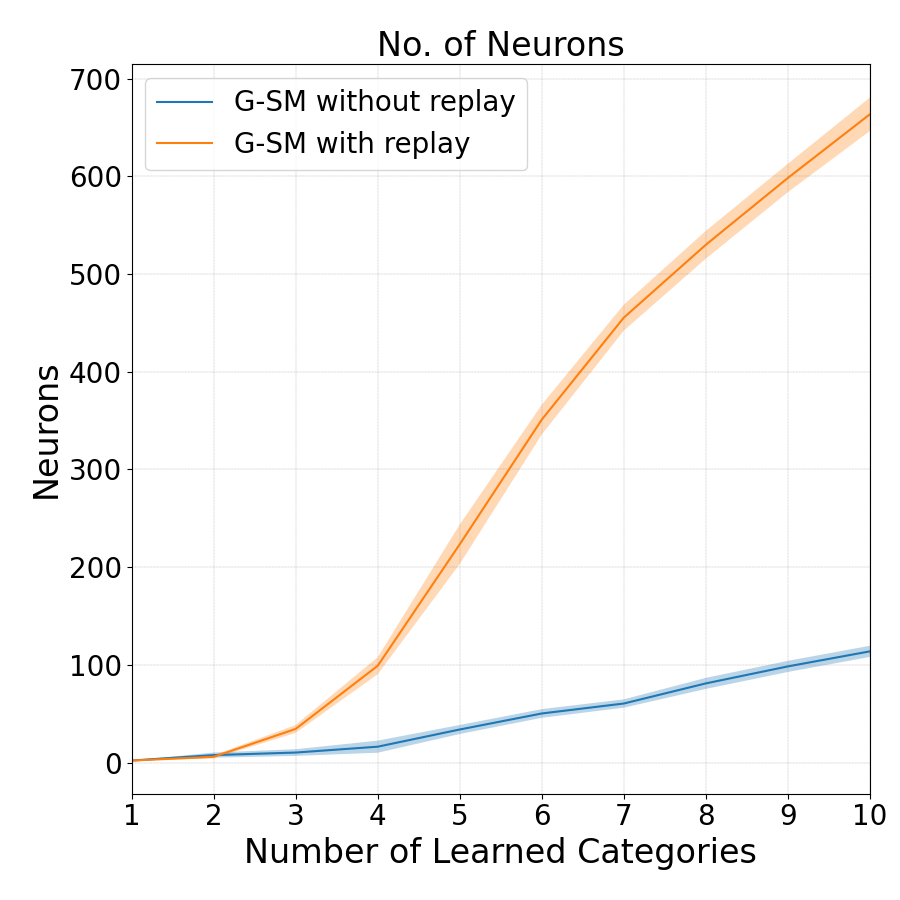}  & \hspace{-5mm}
        \includegraphics[width=0.25\textwidth]{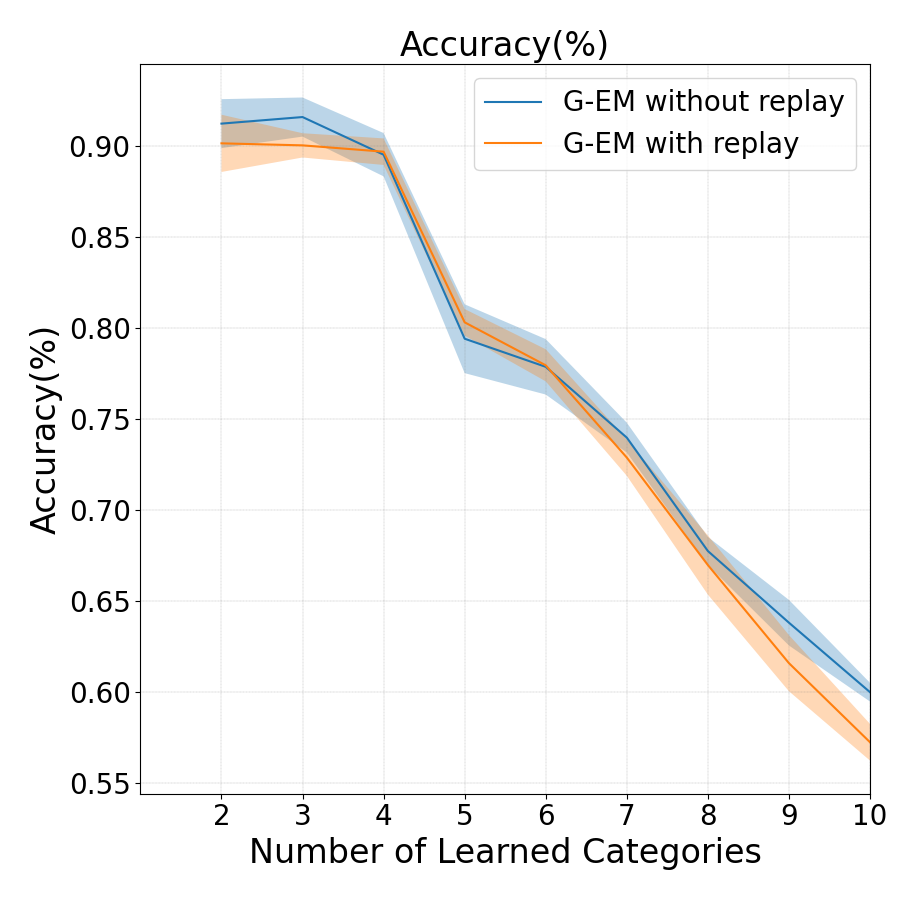} &\hspace{-5mm}
        \includegraphics[width=0.25\textwidth]{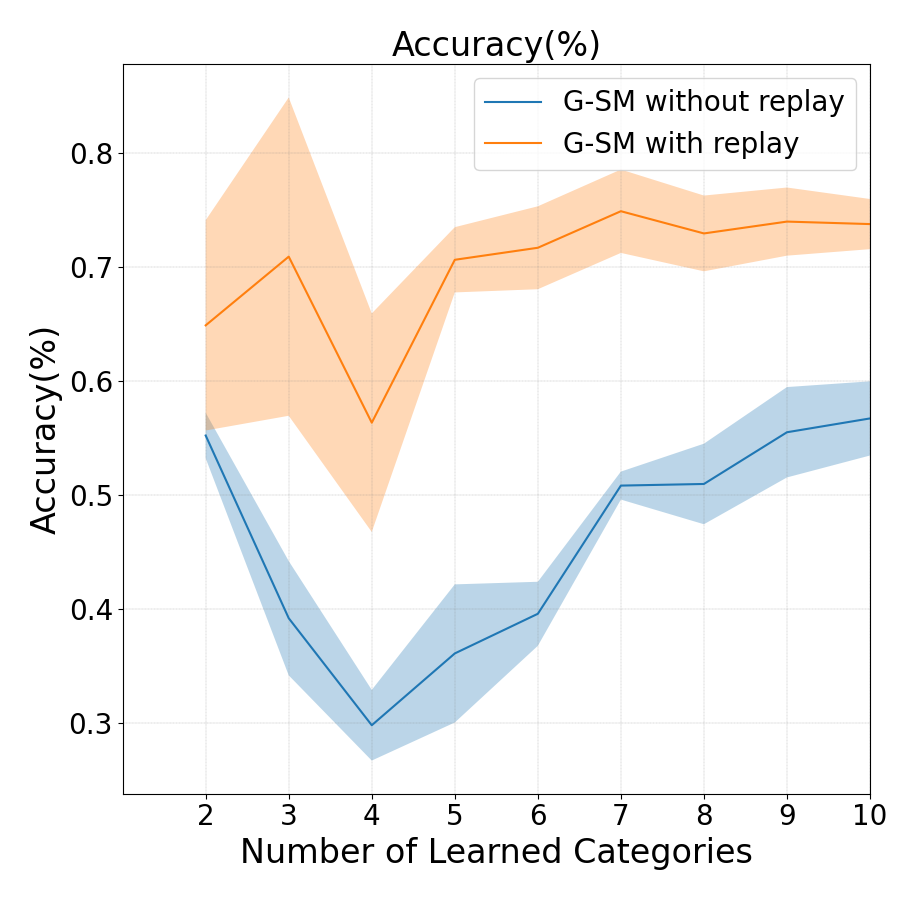}\\
        (a)& (b)& (c)& (d)
    \end{tabular}
    \caption{Comparison results of incremental learning with and without memory replay: (a) Number of neurons in episodic memory, (b) Number of neurons in semantic memory, (c) Episodic memory accuracy, and (d) Semantic memory accuracy. The results are averaged across five learning trials.}
    \label{fig:cmp}
\end{figure}

Figure \ref{fig:9} shows the number of neurons required to learn a certain set of object categories, accuracy as a function of number of learned categories, and the quantization error of the incremental learning scenario over epochs, averaged across five learning trials. 

In batch learning experiments, we observed that the number of required neurons grows exponentially in initial epochs (Fig.~\ref{fig:8}(a)), which indicates high neural activity for the input data distributions. In incremental learning, we observed that the number of required neuron grows (Fig.~\ref{fig:9}(a)) progressively over time in both the networks when a new object category is introduced to the model. We also observed a monotonically increase and decrease in quantization error (Fig.~\ref{fig:9}(c)) in both the networks when a new object category is taught. By increasing the number of categories, the classification task becomes more and more difficult. Therefore, it is expected the overall accuracy in both instance and category decreases as new object categories are introduced. We observed the sudden drop in instance and category accuracy when the \texttt{Staple} object ($2^{nd}$ category) is introduced after the model is trained with the \texttt{Airplane} category. This may be due to that the airplane object has high-level features compared to the staple object, but after learning the staple object we observed the increase in category accuracy. 
%We also noted similar behavior during the learning process of \texttt{Spoon}, \texttt{Pen}, and \texttt{Eraser} objects. 
This clearly shows that the order of object categories has higher sensitivity towards the model performance. Figure \ref{fig:cmp} shows the comparison between the model with and without memory replay in episodic and semantic memory. By comparing the neuron growth between the models, (Fig.~\ref{fig:cmp} (a) and (b)) the memory replay influences the neuron growth in both episodic and semantic memory. In contrast to the batch learning and the model without memory replay, the number of neurons in semantic memory is higher than episodic memory in the model with memory replay. This change in behavior with memory replay is due to the addition of external noise in the dataset. Sometimes adding external noise to the point cloud may change the structure of the object, which results in a completely different shape. Such data samples result in high neuron activity in the GDM networks and during the memory replay phase, and as a consequence, new nodes are added. 

Since there is no periodic reply in the model without MR less neuron growth is observed. When comparing the accuracy results (Fig.~\ref{fig:cmp} (c), and (d)), the instance accuracy of both models remains more or less the same, whereas the memory replay model shows a significant improvement in category accuracy for all the object categories. Based on our experimental results, it can be concluded that the intrinsic memory replay helps to improve the model performance in incremental learning scenarios. In particular, we observed that the periodic replay of temporal trajectories (RNATs), learned from G-EM, could mitigate the problem of catastrophic forgetting.

\begin{wraptable}{r}{0.4\textwidth}
\vspace{-4mm}
\caption{Performance of object recognition on the incremental learning scenario.}
\resizebox{\linewidth}{!}{%
\begin{tabular}{@{}ll@{}}
\toprule
\textbf{Approach}  & \textbf{Accuracy (\%)} \\\midrule
NI - GDM with replay     & {$71.64 \pm 0.13$} \\
NI - GDM without replay  & {$66.00 \pm 0.21$} \\ 
Cummulative \cite{pmlr-v78-lomonaco17a} & {$65.15 \pm 0.66$}\\
LwF \cite{li2017learning} & {$59.42 \pm 2.71$}\\
EWC \cite{kirkpatrick2017overcoming} & {$57.40 \pm 3.80$}\\\hline
NC - GDM with replay     & {$74.07 \pm 0.14$} \\
NC - GDM without replay  & {$56.69 \pm 0.23$} \\ 
Cummulative & {$64.65 \pm 1.04$}\\
LwF & {$27.60 \pm 1.70$}\\
EWC & {$26.22 \pm 1.18$}\\\hline
NIC - GDM with replay    & {$72.41 \pm 0.12$} \\
NIC - GDM without replay & {$65.86 \pm 0.14$} \\ 
Cummulative & {$64.13 \pm 0.88$}\\
LwF & {$28.94 \pm 4.30$}\\
EWC & {$28.31 \pm 4.30$}\\
\bottomrule
{\ul } 
\end{tabular}
}
\vspace{-4mm}
\label{table:cont}
\end{wraptable}
\noindent \textbf{Continuous object recognition:}
We evaluate the incremental learning model with the three continuous object recognition scenarios, proposed by Lomonaco et al. recently \cite{pmlr-v78-lomonaco17a}. The learning tasks include new instances (NI), new categories (NC), and new instances and categories (NIC). In NI settings, the model is initially trained with data of known categories, then, new instances that belong to the known object categories from different acquisition collections are becoming gradually available over time. Hence, the model must be dealt with learning new instances of already known object categories and making correct predictions.

For the NI scenario, the model is initially trained with the first training collection and then, incrementally trained with the remaining $11$ collections. In NC, new classes belong to the different object categories are progressively available over time. Therefore, the model must be dealt with learning new object categories while retaining the knowledge about previously learned categories. For the NC scenario, the model is trained with four training mini-batches. In the first batch, $4$ object categories from all the training collections are included and the remaining three batches include $2$ object categories each. In the NIC setting, samples belong to new instances and categories became available over time, requiring the model to learn new ones while retaining the previously learned ones. NIC training consists of $48$ mini-batches, created from $12$ collections and $10$ object categories. The first batch contains $4$ classes to maximize the categorical representation and the remaining batches include objects from $2$ classes with only one training sequence per class is included.

The classification accuracy of continuous batch learning on the test samples is summarized in Table~\ref{table:cont}. The test set sample contains the samples from all the categories to keep consistency across all three different learning scenarios. Similar to the incremental learning results (shown in Table~\ref{table:inc}), in the NI scenario, NC scenario, and NIC scenario, the model with memory replay achieved a better classification accuracy than the model without memory replay. 
%In contrast, the model without memory replay showed better performances in the NC, and NIC scenarios. 
{Comparison with other continual learning methods, i.e., \cite{lomonaco2017core50} \cite{li2017learning} \cite{kirkpatrick2017overcoming}, are summarized in Table~\ref{table:cont}. By comparing all approaches, it is visible that GDM with memory replay obtained the best results across all the continuous object recognition scenarios.}

{
We performed another round of experiment to measure the model size and inference time of GDM models in three settings: (\textit{i}) batch learning, (\textit{ii}) incremental learning without memory reply, and (\textit{iii}) incremental learning without memory reply. 
The size of networks are averaged across five learning trials with $10$ randomly selected object categories. To measure the average inference time, we randomly select $50$ images and pass them through the network to calculate the time of feed-forward. Then, we average the time difference results. Note that as the inference times are on a millisecond (ms) scale, we just reported the average time. The obtained results are summarized in Table~\ref{tab:computation}.
By comparing all approaches, it is clear that size of episodic (G-EM) and semantic (G-SM) memories in batch learning are significantly larger than that of the incremental learning settings. The underlying reason is that in batch learning setting, samples from all categories are shown to model at each learning epoch, therefore, the number of neural growth increases rapidly (see Fig.~7a). This behaviour results in relatively larger models and longer inference time as compared to incremental learning settings.}
\begin{table}[!h]
\centering
\caption{{Size of trained models and inference time of the proposed approach.}}
\begin{tabular}{lccc}
\toprule
\multicolumn{1}{c}{\multirow{2}{*}{\textbf{Approach}}} & \multicolumn{2}{c}{\textbf{Size (MB)}}  & \multirow{2}{*}{\textbf{Inference Time (ms)}} \\ \cmidrule(lr){2-3}  
\multicolumn{1}{c}{}& \multicolumn{1}{l}{\textbf{G-EM}} & \multicolumn{1}{l}{\textbf{G-SM}} &  \\ \midrule
GDM with Batch Learning                    & 34.78                    & 18.52                    & 97.4                                 \\
GDM with Incremental Learning (without MR) & 06.80                    & 03.52                    & 25.7                                 \\
GDM with Incremental Learning (with MR)    & 14.86                    & 18.74                    & 72.3                                  \\ \bottomrule
\end{tabular}
\label{tab:computation}
\end{table}

\subsubsection{Controlling Connections Removal}
\label{controlled_connections_removal}

\begin{figure}[!t]
% \hspace{-5mm}
    \begin{tabular}{cc}
        \includegraphics[width=0.45\textwidth]{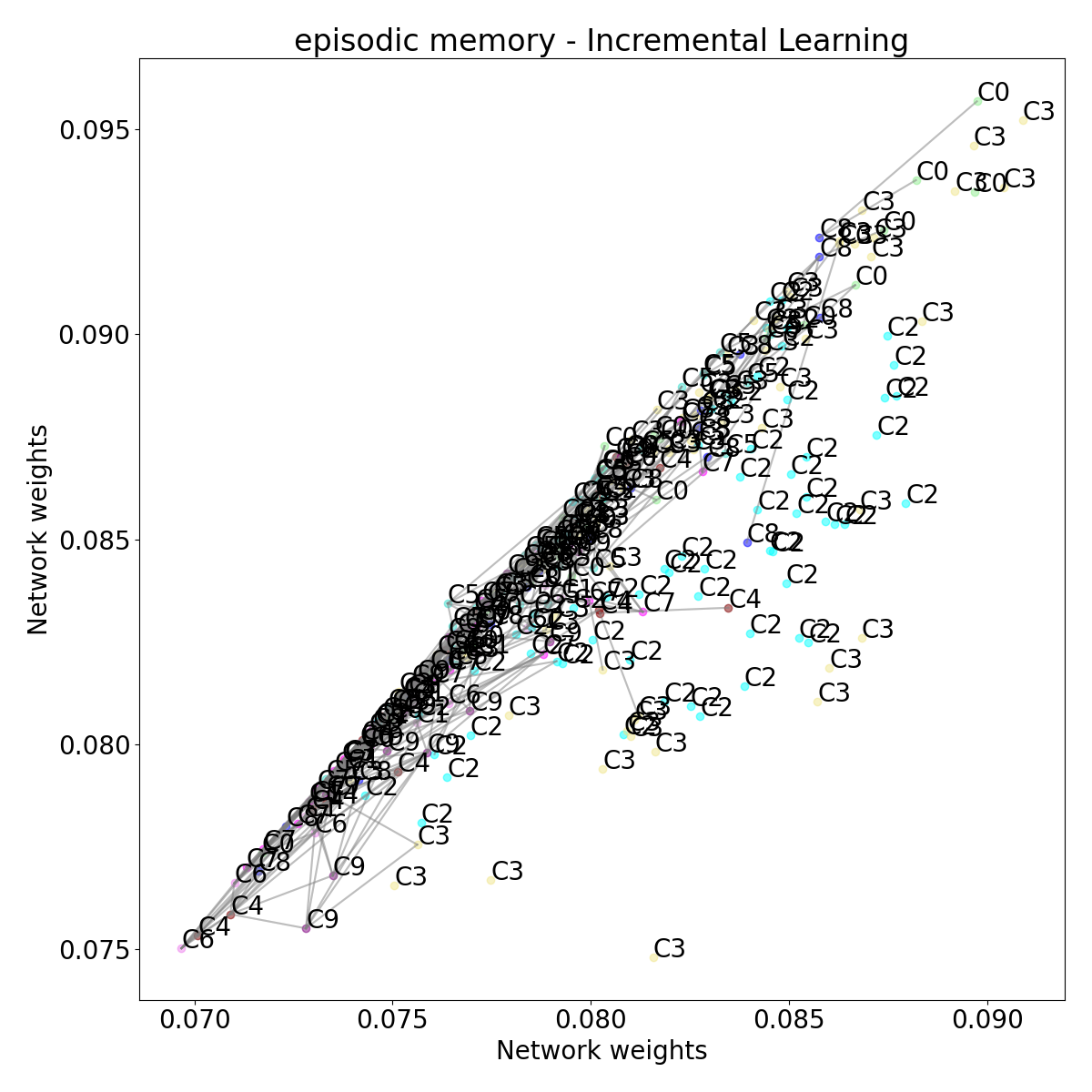} &
        \includegraphics[width=0.45\textwidth]{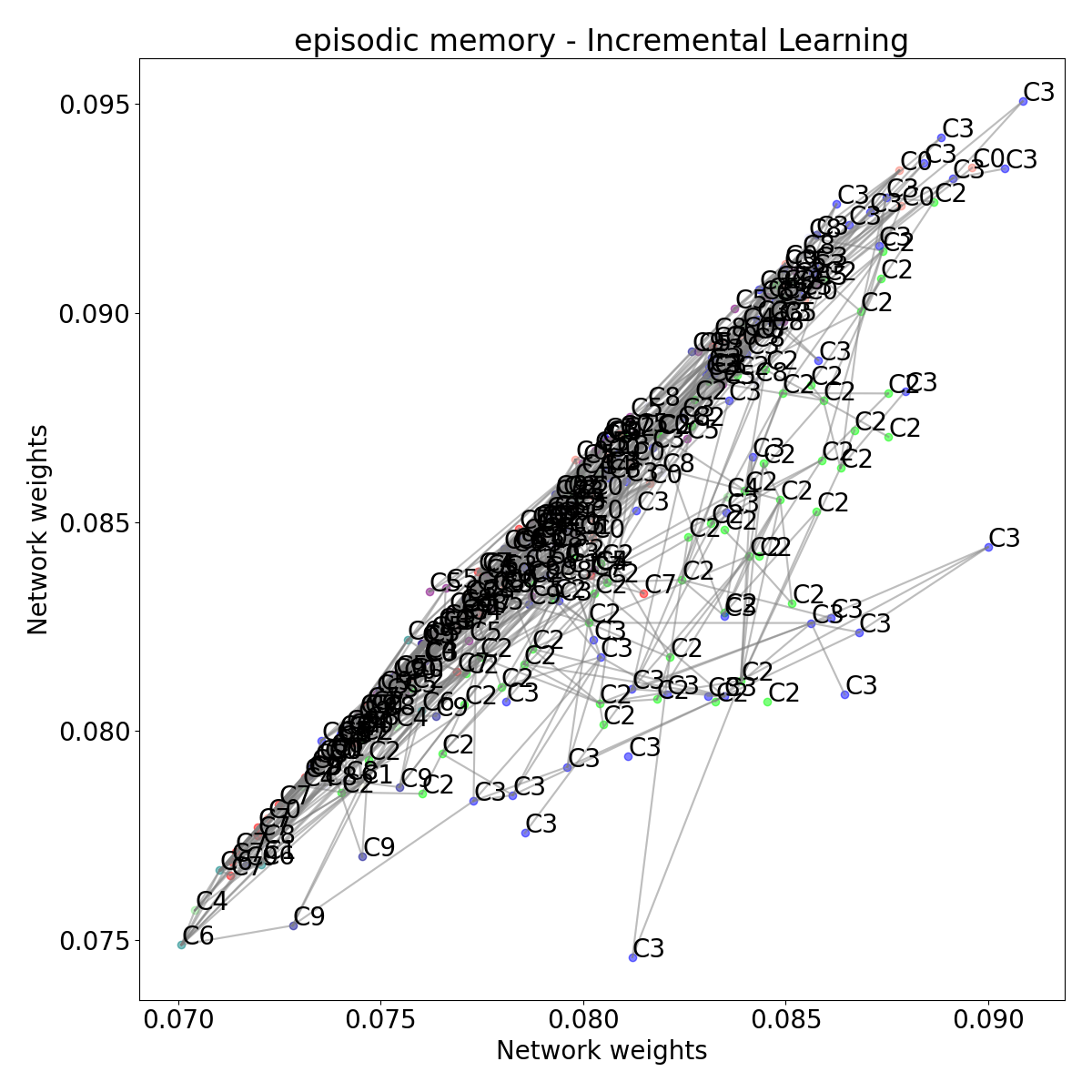}\\
        (a)& (b)
    \end{tabular}
    \caption{The plot of episodic memory with and without controlled connection removal policy: (a) without controlled connections removal, and (b) with controlled connections removal. These plots show the network representation for $10$ categories ($c0 - c9$). The scatter plots are the 2D representation of network weights, where the dimension of network weights (i.e., $256$ dimension) are reduced using principal component analysis.}
    \label{fig:neuron_rm}
\end{figure}

As we discussed earlier, we observed that at the end of an experiment, neurons with good habituation value (i.e., the neurons which represents good knowledge about particular object category) is being removed due to its maximum age criteria and no neighbouring connections. As a consequence, the accuracy of the model dropped in both episodic and semantic memory predictions, which was more severe in episodic memory. This is because the semantic memory neuron growth is regulated (i.e., G-SM adds new neurons only when the BMU label prediction is misclassified with the ground truth data) whereas in the episodic memory, it is unregulated (i.e., the neuron grows whenever new input activity is observed). This results in more number of neurons are getting removed in episodic memory than semantic memory at the end of learning procedure. As an example, Fig.~\ref{fig:neuron_rm}(a) shows the network representation of episodic memory with neighboring connections removed at the end of $10^{th}$ epoch. From this figure, we can see that most of the neurons representing the class $c2$, and $c3$ are left unconnected due to their maximum age despite their knowledge level in the network. To solve this issue, we proposed the controlled edges (connections) removal technique (see section \ref{sec:controlled_removal}). Figure~\ref{fig:neuron_rm} (b) shows the result of the proposed approach. By comparing the obtained results, we can observe the influence of neuron connection removal threshold ($N_T$). In particular, most of unconnected $c2$'s, and $c3$'s neurons, shown in Fig.~\ref{fig:neuron_rm} (a), are retained and adapted well to the input data distribution over time by finding their neighbours and moving close to their associated neighbours (see Fig.~\ref{fig:neuron_rm} (b)). 

\begin{figure}[!b]
\hspace{-5mm}
    \begin{tabular}{cccc}
        \includegraphics[width=0.25\textwidth]{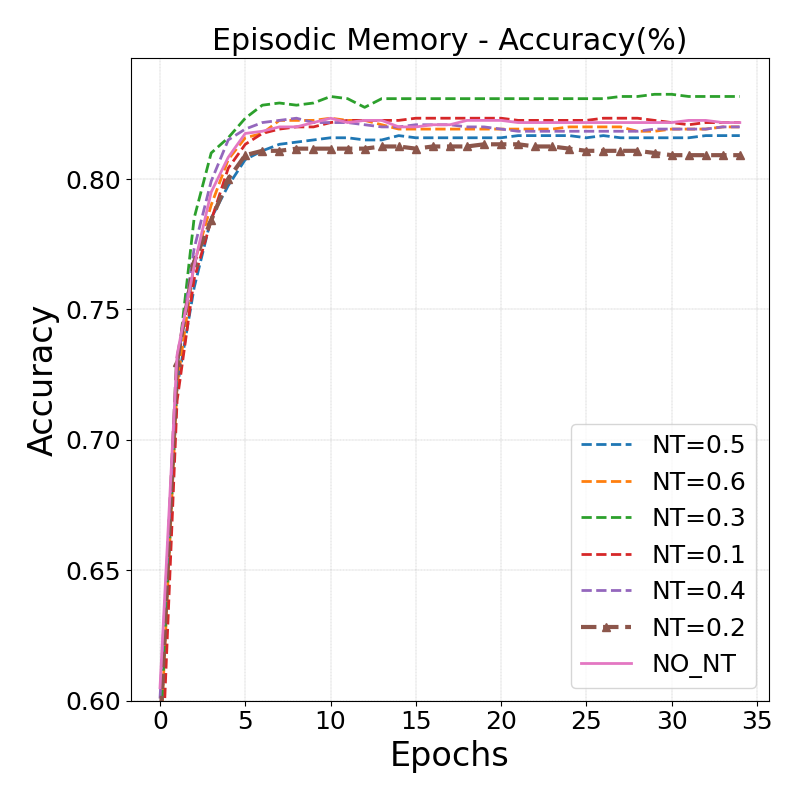} &\hspace{-5mm}
        \includegraphics[width=0.25\textwidth]{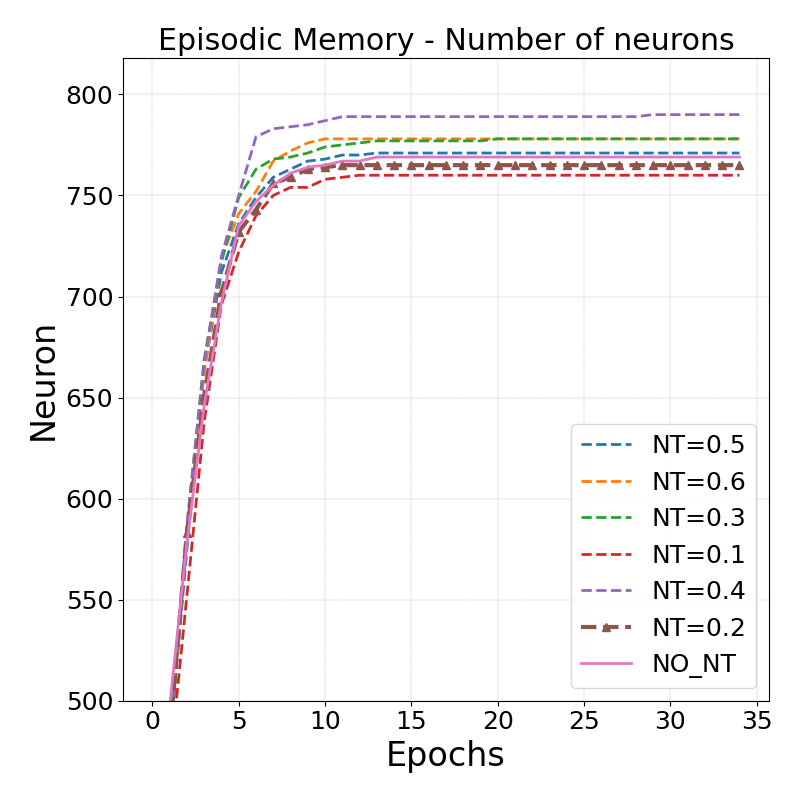}  & \hspace{-5mm}
        \includegraphics[width=0.25\textwidth]{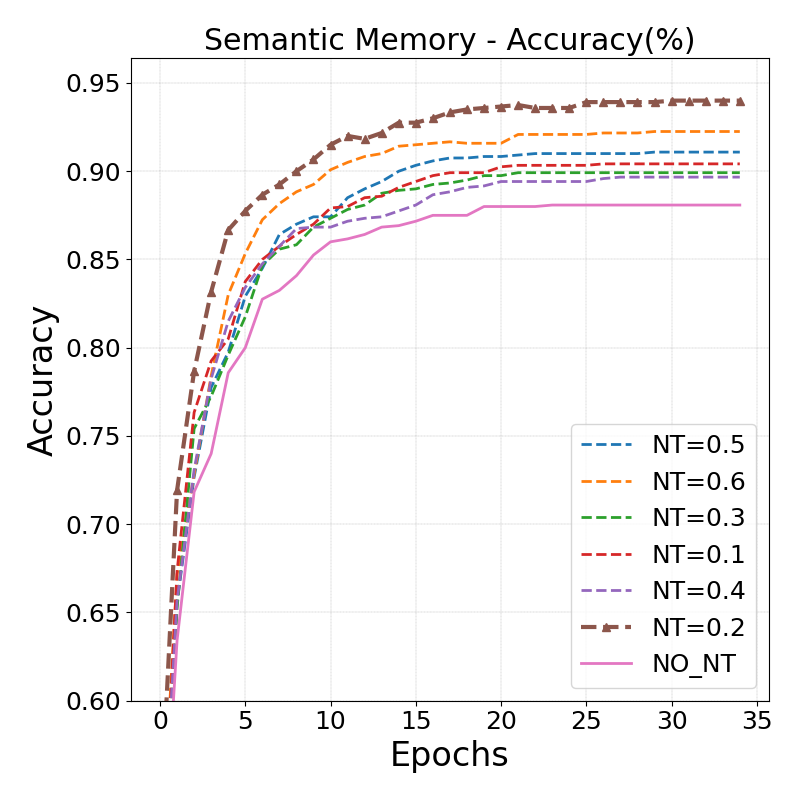} &\hspace{-5mm}
        \includegraphics[width=0.25\textwidth]{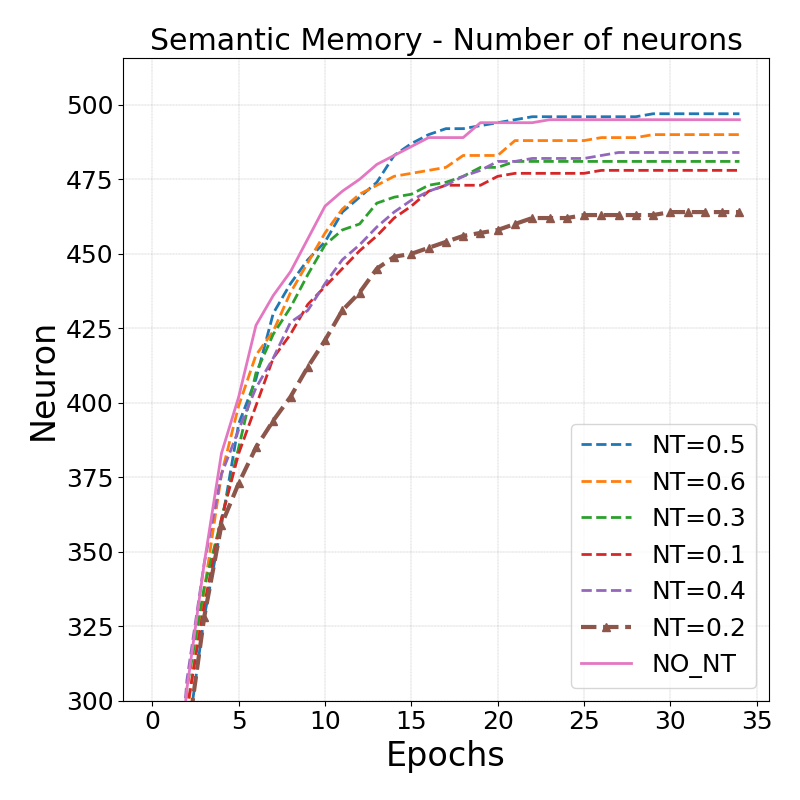}\\
        (a)& (b)& (c)& (d)
    \end{tabular}
    \caption{The effect of controlled edges (connections) removal on different $N_T$ threshold values on: (a) episodic memory accuracy, (b) neurons growth in episodic memory, (c) semantic memory accuracy, and (d) neurons growth in semantic memory over different learning epochs. The dashed plot with triangles ($N_T$ = 0.2) shows the best overall performance.}
    \label{fig:rm_eval}
\end{figure}

We performed a set of experiments to evaluate the effect of $N_T \in \{0.0, 0.1$, \dots, $0.6\}$ on the accuracy and number of required neurons in episodic memory and semantic memory. The obtained results are shown in Fig.~\ref{fig:rm_eval}. 

The dotted lines in Fig.~\ref{fig:rm_eval} indicates the result obtained with varying $N_T$, while the pink line indicates the result of $N_T=1$ (NO\_NT). The best result was obtained with $N_T=0.2$ (shown by dotted line with triangle indicates). By comparing the effect of $N_T$ in the semantic memory (Fig.~\ref{fig:rm_eval} \textit{(c) and (d)}), the NO\_NT model (pink line) shows the low category-level prediction accuracy and high neurons growth, whereas the $N_T$ helps to improves the prediction accuracy overall for all the different values. Regrading the semantic memory (Fig.~\ref{fig:rm_eval} \textit{(c) and (d)}), $N_T=0.2$ shows the best overall category-level accuracies when comparing with other $N_T$ values.

%Regarding the semantic memory (Fig.~\ref{fig:rm_eval} \textit{bottom-row}), $N_T=0.2$ and $N_T=0.3$ show the best instance-level accuracies when comparing with other values. 

In contrast to the semantic memory, $N_T=0.3$ gives the high instance-level accuracy when compared with the models trained with other $N_T$ values. Which is $2.27 \%$ higher than the $N_T=0.2$, this is due to that $N_T=0.3$ holds more neurons than the $N_T=0.2$. But when comparing the performance of $N_T=0.3$ and $N_T=0.2$ at the category-level (i.e., at semantic memory), the model with $N_T=0.3$ shows $4.09 \%$ less accuracy performance than the model with $N_T=0.2$. This clearly shows that neurons in model with $N_T=0.2$ represents the best knowledge about the input data distribution. 
When comparing the overall performance in both semantic and episodic, even though the model with $N_T=0.3$ excels at performance on episodic memory, the model with $N_T$ value of $0.2$ achieved the overall best performance at semantic memory and holds good performance at episodic memory. Hence, we used $N_T$ value of $0.2$ for training the GDM model on all the batch and incremental learning scenarios. It should be noted that the results shown in Fig.~\ref{fig:rm_eval} are obtained using the GDM model trained on batch learning with no temporal context, as it trains faster compared to other models. 

\subsection{Robot Experiments}
\begin{wrapfigure}{r}{0.4\textwidth}
    \vspace{-3mm}
    \centering
    \includegraphics[width=\linewidth]{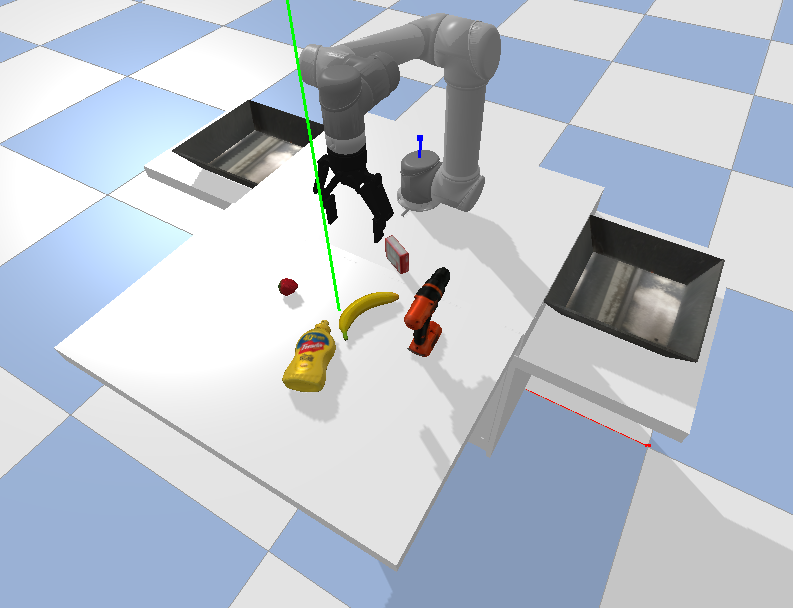}
    \caption{ Our experimental setup consists of a table, two baskets, a URe5 robotic arm, and objects from YCB dataset~\cite{doi:10.1177/0278364917700714}. The green line indicates the camera line of sight.}
    \label{fig:sim_exp}
    \vspace{-5mm}
\end{wrapfigure}

\begin{figure}[!b]
    \centering
    \includegraphics[width=\linewidth]{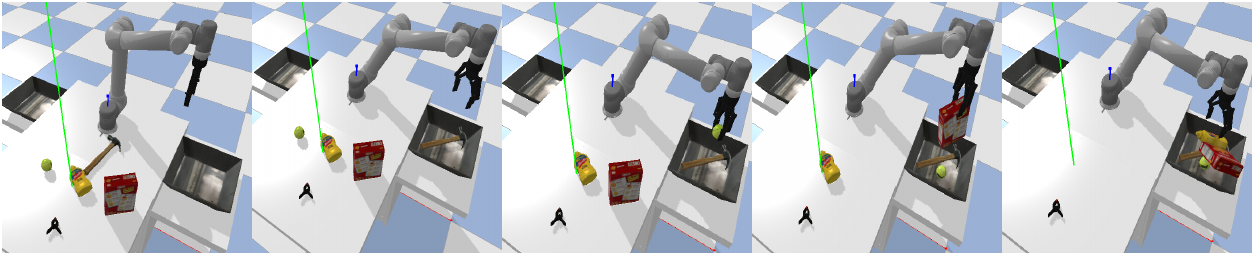}\\
    \includegraphics[width=\linewidth]{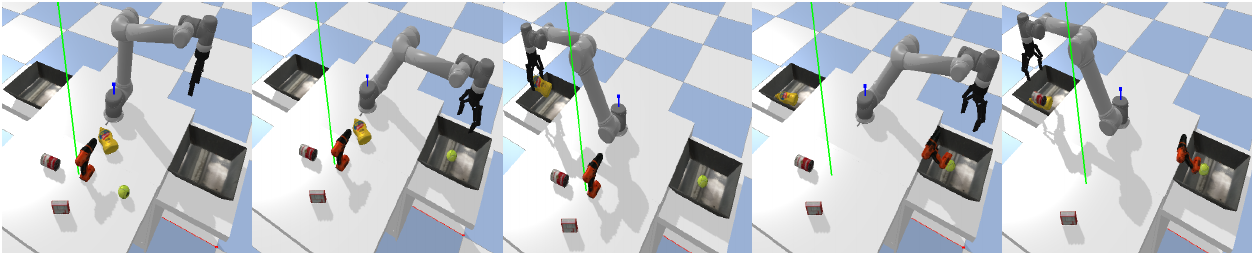} 
    \caption{Sequence of snapshots of simulated robot experiments: (\textit{top-row}) in a pick and place scenario, each object on the table needs to be grasped and placed in left basket; (\textit{bottom-row}) in a packing scenario, a set of objects (i.e., \texttt{mustard\_bottle}, and \texttt{tomato\_soup\_can}) needs to be placed in the right basket and the remaining objects needs to be placed in left basket. To accomplish this task, the robot needs to recognize all objects first and then put them into the desired baskets.}
    \label{fig:robot_exp}
\end{figure}
We evaluated the performance of the proposed approach in a simulated robot environment. Our experimental setup is shown in Fig.~\ref{fig:sim_exp}. We used the UR5e robotic arm as a manipulator and an RGB-D camera to perceive the environment. We imported $15$ simulated objects from the YCB dataset \cite{doi:10.1177/0278364917700714} to test the proposed approach. 

In this round of evaluation, we considered two tasks: (\textit{i}) pick and place in the context of clear the table, and (\textit{ii}) a packing scenario, where the robot should organize objects in specific baskets. In all the experiments, several objects will randomly placed in front of the robot. To segment the objects from each other, we developed a contours-based bounding box detection to detect the object boundaries. The performance of the experiments is assessed by calculating the success rate, i.e. $\frac{\operatorname{number~of~success}}{\operatorname{total~number~of~attempts}}$. {For the packed scenario, in addition to the success rate, we report the average percentage of objects removed from the workspace. An experiment is continued until either all objects get removed from the workspace, or three failures occurred consecutively. Note, the experiments will be counted as a success only when the target objects are placed inside the desire baskets. We compared our approach with two state-of-the-art grasping approaches, including GGCNN~\cite{doi:10.1177/0278364919859066} and GR-ConvNet~\cite{kumra2021antipodal}}. 

\noindent \textbf{Pick and place experiments:} This set of experiments is set up in a way that the robot needs to learn and recognize the object first before grasping and manipulating it to the basket. If the given test object is \texttt{unknown} to the robot, it has to learn the object category first, based on the teacher input (see Fig.~\ref{fig:1}). In particular, we included a recognition task in this experiment to ensure that the GDM model can learn new object categories in a lifelong setting and can recognize them without catastrophic forgetting. If the model can not recognize the object correctly, simulate teacher provides corrective feedback to the GDM model as shown in see Fig.~\ref{fig:1}.

In each experiment, we randomly placed five test objects on the table, and tested each simulated object $50$ times. A sequence of snapshots of a sample experiment is shown in Fig.~\ref{fig:robot_exp} (\textit{top-row}). We achieved $80.27\%$ success rate (i.e. $602$ success out of $750$ attempts). In pick and place experiment, the most failure cases were due object slipped from the gripper, poor grasp quality score at specific object pose, and inaccurate calibration of camera pose relative to the arm leading to collision between the gripper and the target object.

\begin{wrapfigure}{r}{0.4\textwidth}
    \vspace{-3mm}
    \centering
    \includegraphics[width=\linewidth]{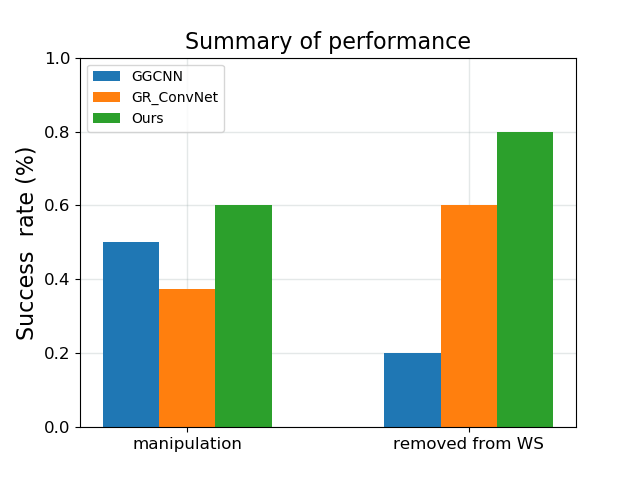}
    \caption{{Comparison of the proposed approach with GGCNN and GR-ConvNet.}}
    \label{fig:comparison}
    \vspace{-5mm}
\end{wrapfigure}
\noindent \textbf{Packing experiments:} In these experiments, two baskets (named right\_basket, and left\_basket) are placed in the environment, where the instances of the selected categories (based on the user input e.g., \texttt{Scissors} and \texttt{Mug}) needs to be packed in the right\_basket and the remaining objects should be placed in the left\_basket. If none of the object categories are selected all the objects need to be placed in the left\_basket. At the beginning of each experiment, we have randomly shuffled object categories, and five objects are randomly selected and placed on top of the table. Figure \ref{fig:robot_exp} (\textit{bottom-row}) shows a sequence of snapshots for a packing scenario, where the robot needs to place the target object (i.e., \texttt{mustard bottle, and tomato\_soup\_can}) to the \textit{right\_basket} and the remaining objects to the \textit{left\_basket}.

The success will be counted only when the objects are placed in the desire baskets. We conducted the packing experiment for $50$ times. {Since the experiment was aborted after three consecutive unsuccessful grasps, or when the robot was unable to find high quality grasp points, not always all objects were cleared from the workspace (WS). The obtained results are depicted in Fig.~\ref{fig:comparison}.}   

We achieved the grasp success rate of $59.20\%$ (i.e., $148$ success out of $250$ attempts). {By comparing the obtained results, it is clear the proposed approach clearly outperformed both GGCNN and GR-ConNet. More specifically, in the case of manipulation success rate, our approach achieved $59.2\%$ success rate which was $~9.2\%$ and $22.5\%$ better than GGCNN, and GR-ConvNet, respectively. In terms of the average percentage of objects removed from the workspace, the proposed method achieved $80\%$, whereas GGCNN and GR-ConvNet achieved $20\%$ and $60\%$ respectively.} 

{A possible explanation for the performance differences is due to the different input modalities. Whereas GG-CNN relies on depth only information, GR-ConvNet and our approach use RGB and depth data that can help to distinguish objects from each other and also from the table. More specifically, we observed that it was hard for the robot to perceive small flat objects (e.g., scissors) when only relying on depth data. Furthermore, the difference in performance between our approach, GR-ConvNet and GG-CNN could be explained by the extra residual layers in between the encoder and decoder part in our approach and in GR-ConvNet’s architecture. Such architectures assist the network to learn representative features from the input data.}

We observed that, for some of the experiments, even though the model predicts the correct object labels, the robot couldn't place the object to target basket due to an inaccurate bounding box detection. In packing scenarios, we observed that most of the failures happened due to (\textit{i}) the object attached to the gripper get slipped while reaching the basket, (\textit{ii}) misclassification of the target object, (\textit{iii}) the collision between the object and the robot gripper, and (\textit{iv}) incorrect bounding box estimations. A video of these experiments is available online at: \url{https://youtu.be/AaWppzGeh9E}.

%%%%%%%
\section{Conclusion}
\label{conclusion_future_work}
In this paper, we presented a model using a growing dual-memory network (GDM) and autoencoder to simultaneously handle object recognition and grasping tasks. The experimental results by GDM on the batch and incremental learning showed that the GDM model was able to learn about new objects in both instance- and category-level over time. We also addressed the problem of catastrophic forgetting by the intrinsic memory reply using RNAT's based on the learned temporal knowledge. To demonstrate the performance of the model in handling object recognition and grasping tasks simultaneously, we performed a set of robot experiments in the context of pick and place and packing scenarios. Results showed that the model was able to recognize the target objects accurately, and predict pixel-wise grasp configuration for performing manipulation task. 

%In future work, we like to further investigate the choice of hyper-parameters to improve the performance of the GDM learning. In a real-world setting, the number of samples to learn the particular object category is limited, so in our further work GDM learning setting which helps to maximize the classification performance with fewer samples needs to be investigated.

In this paper, all training and testing samples are synthesized from the simulated 3D object. Therefore, as a future work, we would like to fine-tune and test the proposed approach using real-data to investigate the possibility of sim2real transfer learning in the context of lifelong learning. As another direction, we would like to investigate the possibility of improving the performance by utilizing networks that directly use three-dimensional data instead of images (e.g., Res-U-net \cite{li2020learning}).   

\bibliography{example}  % .bib

%%%%%%%

%===============================================================================

% The maximum paper length is 8 pages excluding references and acknowledgements, and 10 pages including references and acknowledgements

\clearpage
% The acknowledgments are automatically included only in the final and preprint versions of the paper.
%\acknowledgments{If a paper is accepted, the final camera-ready version will (and probably should) include acknowledgments. All acknowledgments go at the end of the paper, including thanks to reviewers who gave useful comments, to colleagues who contributed to the ideas, and to funding agencies and corporate sponsors that provided financial support.}

\appendix
\appendixpage

\section{GDM Hyperparameter selection}
\label{sec:appendix_b}

\subsection{Batch Learning}
The hyper-parameters used during the batch learning are summarized in Table \ref{tablehp}.

\begin{table}[!th]
\centering
\caption{Hyper-parameter settings for G-EM and G-SM networks in batch learning.}
\vspace{2mm}
\begin{tabular}{@{}lllll@{}}
\toprule
\textbf{Hyperparameter}  & \textbf{Values} & \\ \midrule
Insertion thresholds     & $a^{EM}_T = 0.7$, $a^{SM}_T = 0.8$                &  \\
Global context           & $\beta = 0.5$               &  \\
Learning rates           & $\epsilon_b = 0.3, \epsilon_n = 0.003, \epsilon_c = 0.001$               & \\
Habituation threshold    & $h_T = 0.1$                &  \\
Habituation function     & $\tau_b = 0.3, \tau_i = 0.1, \kappa = 1.05$                &  \\
Neuron removal threshold & $N_T = 0.2$                &  \\ 
Labelling                & $\delta^+ = 1, \delta^- = 0.1$                &  \\
Context descriptors      & $\alpha_1 = 0.63, \alpha_2=0.234, \alpha_3=0.086$ &  \\ \bottomrule
{\ul }                   &                 & 
\end{tabular}
\label{tablehp}
\end{table}

\subsection{Incremental Learning}
The hyper-parameter settings that lead to the best performance for the incremental learning are shown in Table \ref{table:hp_inc}. 

\begin{table}[!th]
\caption{Hyper-parameter settings for G-EM and G-SM networks in incremental learning.}
\vspace{2mm}
\centering
\begin{tabular}{@{}lllll@{}}
\toprule
\textbf{Hyperparameter}  & \textbf{Values} & \\ \midrule
Insertion thresholds     & $a^{EM}_T = 0.5, a^{SM}_T = 0.7$                &  \\
Global context           & $\beta = 0.4$                &  \\
Learning rates           & $\epsilon_b = 0.5, \epsilon_n = 0.005, \epsilon_c = 0.001$               & \\
Habituation threshold    & $h_T = 0.1$                &  \\
Habituation function     & $\tau_b = 0.3, \tau_i = 0.1, \kappa = 1.05$                &  \\
Neuron removal threshold & $N_T = 0.2$                &  \\ 
Labelling                & $\delta^+ = 1, \delta^- = 0.1$                &  \\
Context descriptors      & $\alpha_1 = 0.63, \alpha_2=0.234, \alpha_3=0.086$ &  \\ \bottomrule
{\ul }                   &                 & 
\end{tabular}

\label{table:hp_inc}
\end{table}

%===============================================================================
\end{document}